\documentclass[letterpaper, 10 pt, conference]{ieeeconf} 

\IEEEoverridecommandlockouts                            
\usepackage{graphics} % for pdf, bitmapped graphics files
\usepackage{epsfig} % for postscript graphics files
\usepackage{mathptmx} % assumes new font selection scheme installed
\usepackage{times} % assumes new font selection scheme installed
\usepackage{amsmath} % assumes amsmath package installed
\usepackage{amssymb}  % assumes amsmath package installed
\usepackage{amssymb}
\usepackage{graphicx}
\usepackage{amsmath}
\usepackage{algorithm}
\usepackage{algorithmicx}
\usepackage{algpseudocode}
\usepackage{indentfirst}
\usepackage{subfigure}
\usepackage{cite}
\usepackage{color}

\begin{document}

\title{\LARGE \bf
Multi-robot Cooperative Pursuit via Potential Field-Enhanced Reinforcement Learning}

\author{ Zheng Zhang, Xiaohan Wang, Qingrui Zhang, and Tianjiang Hu$^*$
\thanks{All authors are with  with Machine Intelligence and Collective Robotics (MICRO) Lab, Sun Yat-sen University, Guangzhou 510725, China 
}
\thanks{\tt\small zhangzh363@mail2.sysu.edu.cn; }
\thanks{\tt\small 
wangxh258@mail2.sysu.edu.cn; }
\thanks{\tt\small
zhangqr9@mail.sysu.edu.cn;}
\thanks{\tt\small hutj3@mail.sysu.edu.cn}
\thanks{This paper is supported by the National Nature Science Foundation of China under Grant 62103451}
}

\maketitle
\thispagestyle{empty}
\pagestyle{empty}

%%%%%%%%%%%%%%%%%%%%%%%%%%%%%%%%%%%%%%%%%%%%%%%%%%%%%%%%%%%%%%%%%%
\begin{abstract}
It is of great challenge, though promising, to coordinate collective robots for hunting an evader in a decentralized manner purely in light of local observations. In this paper, this challenge is addressed by a novel hybrid cooperative pursuit algorithm that combines reinforcement learning with the artificial potential field method. In the proposed algorithm, decentralized deep reinforcement learning is employed to learn cooperative pursuit policies that are adaptive to dynamic environments. The artificial potential field method is integrated into the learning process as predefined rules to improve the data efficiency and generalization ability. It is shown by numerical simulations that the proposed hybrid design outperforms the pursuit policies either learned from vanilla reinforcement learning or designed by the  potential field method. Furthermore, experiments are conducted by transferring the learned pursuit policies into real-world mobile robots. Experimental results demonstrate  the feasibility and potential of the proposed algorithm in learning multiple cooperative pursuit strategies.
\end{abstract}

\IEEEpeerreviewmaketitle

%%%%%%%%%%%%%%%%%%%%%%%%%%%%%%%%%%%%%%%%%%%%%%%%%%%%%%%%%%%%%%%%%%
\section{Introduction}

Multi-robotic cooperation has received extensive research attention in diverse domains due to its efficiency and capability in conducting complex missions \cite{bayindir2016review}, e.g., cooperative surveillance \cite{castanedo2010data}, search and rescue \cite{allouche2010multi}, and air combat \cite{sun2021multi}, etc. As a typical application scenario, the cooperative pursuit aims to coordinate multiple robots for capturing one or multiple evaders \cite{chung2011search} in a confined environment with obstacles. One of the fundamental challenges is the cooperation among robots in capturing the evader safely.
% One of the fundamental challenges is the development of a cooperative algorithm allowing multiple robots collaborate with one another efficiently and safely in capturing the evader in diverse scenarios.

A straightforward solution to cooperative pursuit is based on a centralized architecture in which actions of  robots are produced  all together by one central computation module \cite{lavalle2001visibility,gerkey2006visibility,zhou2016coopeÒrative,shah2019multi}. However, centralized methods are troubled with ‘the curse of dimensionality’, as the computation cost grows exponentially with the increase of the total number of robots. Hence,  centralized methods are difficult to scale up to a multi-robot system with a large group size. Centralized methods are also vulnerable to communication failures or delays, as the joint actions computed by the central module are sent to robots in a real-time manner.  On the contrary, a decentralized algorithm can distribute the computation burden to all robots in a group, so it is computationally lightweight. In decentralized algorithms, decisions are made based on local observations, which is preferable to large-scale multi-robot systems  \cite{long2018towards}.  

% Since communication between robots may be unreliable in the adversarial environment, we suppose a group of non-communicating robots make decisions based on only local observations. 

In this paper, we propose a decentralized cooperative pursuit algorithm that combines deep reinforcement learning (RL) and the artificial potential field (APF) method. In the proposed design, the last layer of the policy network is designed manually using APF to improve the data efficiency and generalization ability.  Since robots are assumed to have homogeneous dynamics, their policy networks share the same structure and parameters to improve the search efficiency \cite{gupta2017cooperative}. Following the idea in \cite{huttenrauch2019deep}, the mean embedding of distributions is introduced to resolve the varying size issue of the local observations in multi-robot settings. Together with a virtual obstacle mechanism, the wall following rules are borrowed from \cite{borenstein1989real,yun1997wall} to prevent the robots from being trapped in local minima \cite{ge2002dynamic}. The overall contributions of this paper are summarized as follows.
\begin{enumerate}
    \item A novel decentralized pursuit algorithm is proposed to learn cooperative pursuit strategies efficiently that can leverage the merits of both deep RL and APF. 
    \item A modified version of wall following rules and a virtual obstacle mechanism are designed to overcome the local minima issues in APF methods.
    \item The learned pursuit policy is deployed in real-world robots to show the efficiency of the proposed design.
\end{enumerate}

The rest of this paper is organized as follows. In Section \ref{sec:related work}, the related works are summarized. Preliminaries are provided in Section \ref{sec:preliminary}. Section \ref{sec:acp} presents the implementation details of the proposed algorithm. Numerical simulations and experimental results are given in Section \ref{sec:result}. Finally, conclusions and future works are available in Section \ref{sec:conclusion}.
%%%%%%%%%%%%%%%%%%%%%%%%%%%%%%%%%%%%%%%%%%%%%%%%%%%%%%%%%%%%%%%%%%
\section{Related Works}\label{sec:related work}

Many decentralized strategies for cooperative pursuit belong to force-based methods, aiming to design the mathematical form of several forces to guide pursuers' move  \cite{angelani2012collective,khatib1986real}. For example, Janosov et al. assumed that each pursuer is attracted by the evader and repulsed by their teammates \cite{janosov2017group}. However, most force-based methods stress more on greedy chasing of evaders and collision avoidance, instead of cooperation. In \cite{escobedo2014group}, Escobedo advocated time-depending parameters in the aforementioned formulas are a necessity for cooperation emerging. Muro et al. demonstrated several cooperative strategies, e.g. encirclement, ambushing, and relay running, by designing a variable inter-individual repulsion depending on the distance from the evader \cite{muro2011wolf}. In \cite{fang2020cooperative}, a surrounding force, which depends on the distance from neighbors, is introduced to evenly distribute pursuers on the encirclement to the evader. Although cooperation is achieved to some extent in the above works, it is extremely vulnerable to escape policies and initial conditions. Once the evader becomes more intelligent or the task environment becomes more complex, those cooperative strategies tend to collapse.

Another promising strategy for decentralized cooperative pursuit is via deep RL that learns diverse pursuit strategies by interacting with environments \cite{sutton2018reinforcement}. In comparison with force-based methods, deep RL is possible to obtain various abilities, some of which are difficult to hard-code via explicit rules, in the service of maximizing their rewards \cite{silver2021reward}. Wang et al. applied MADDPG to learn complicated cooperative strategies to chase an evader, e.g. surrounding and pushing the evader to the wall \cite{wang2020cooperative}. Gupta employed TRPO with parameter sharing to learn a pursuit policy that coordinates multiple pursuers to attack the evader simultaneously \cite{gupta2017cooperative}, proving deep RL outperforms Joint Equilibrium search for policies \cite{nair2003taming}. In \cite{de2021decentralized},  De Souza et al. injected domain-specific knowledge into deep RL by a novel reward function, which awards the pursuers when they spread around the evader. As a result, encirclement behaviors emerge to facilitate cooperation between pursuers. However, deep RL is notorious for its sample inefficiency and poor generalization ability \cite{rlblogpost}. Hence,  it is still an open issue to efficiently learn cooperative strategies that can generalize to different situations.
%%%%%%%%%%%%%%%%%%%%%%%%%%%%%%%%%%
\section{Preliminaries}\label{sec:preliminary}

\subsection{Problem Formulation}\label{subsec:problem}
\begin{figure}
    \centering
    \includegraphics[width=0.4\textwidth]{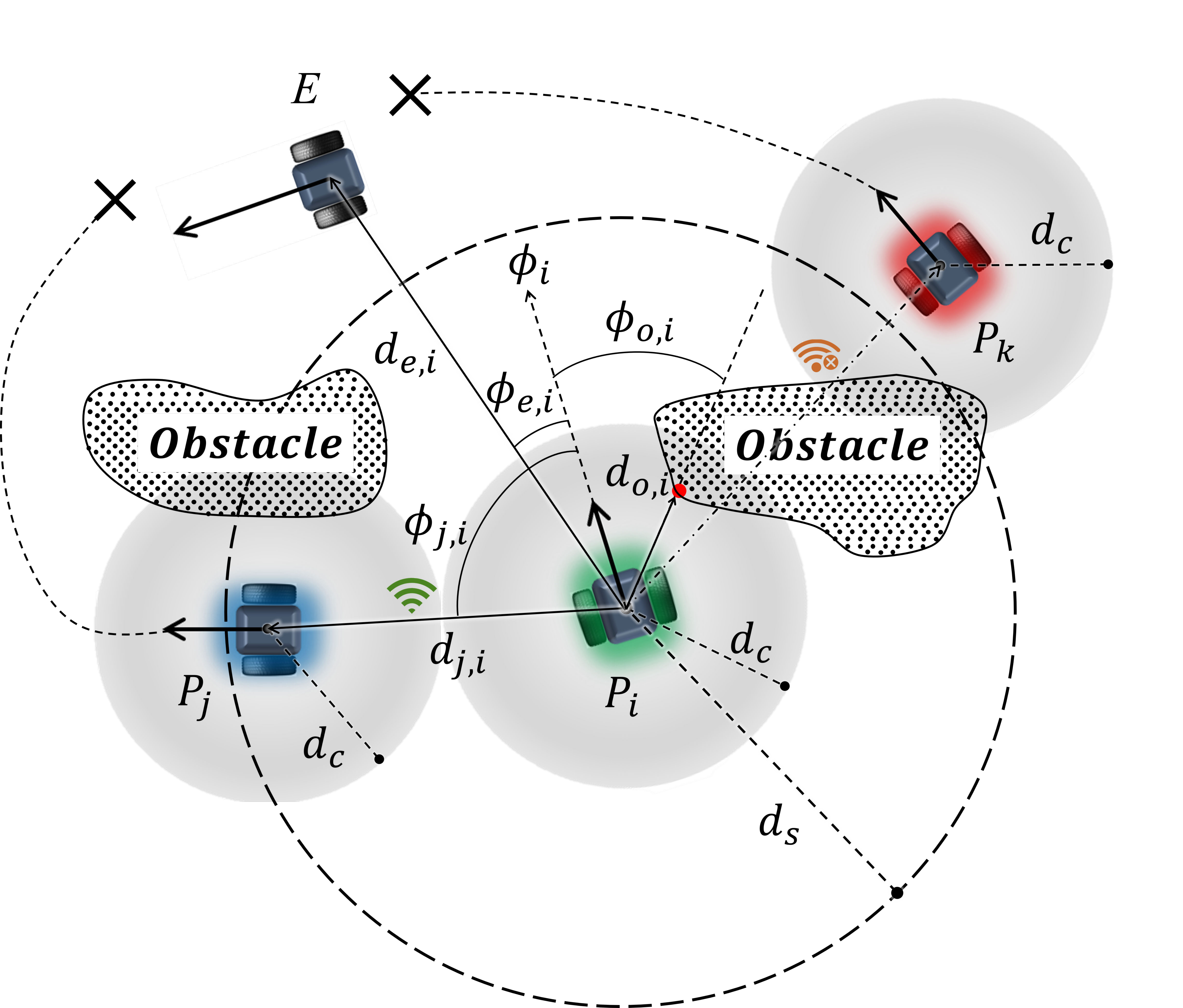}
    \caption{The observation space of pursuers ($E$ is the evader, while $P$ is the pursuers). The pursuer $P_i$ can not observe $P_k$ due to the long distance in between. The pursuer can perceive the distance and bearing of the nearest point related to obstacles.}
    \label{fig:state space}
\end{figure}

In this paper, the cooperative pursuit problem is defined for $N$ robots pursuing a faster evader in an environment with obstacles. The evader will be \emph{captured}  by a pursuer if the distance between them is less than $d_c$. The mission is complete, once the evader is captured successfully by all pursuers. The following assumptions are further made.
\begin{enumerate}
    \item The pursuing robots (pursuers) and evading robot (evader) have the radius of $d_p$ and $d_e$, respectively;
    \item All pursuers are homogeneous and they can only observe neighbours in a distance of $d_s$ as shown in Figure \ref{fig:state space}.
    \item The pursuers and evader move at a constant speed $v_p$ and $v_e$, respectively, and $v_e>v_p$;
    \item The evader position is available to all pursuers.
\end{enumerate}

 As shown in Figure \ref{fig:state space}, the local observation of the $i$-th pursuer contains the distance from the nearest obstacle $d_{o,i}$, the evader $d_{e,i}$, the teammates $d_{1,i},…, d_{M,i}$, and the corresponding bearing $\phi_{o,i}, \phi_{e,i}, \phi_{1,i},…,\phi_{M,i}$, where $M$ is the number of observed teammates.

\subsection{Deep Reinforcement Learning}
 The cooperative pursuit problem can be formulated as a standard Markov Decision Process  described by a 6-tuple $(\mathcal{S},\mathcal{A},\mathcal{P},\mathcal{R},\gamma)$, where $\mathcal{S}$ is the state space, $\mathcal{A}$ is the action space, $\mathcal{P}$ is the state-transition model, $\mathcal{R}$ is the reward function, $\gamma$ is the discount factor. The objective of deep RL is to learn an optimal policy $\pi^*(\boldsymbol{a}|\boldsymbol{s})$ that maximizes the expectation of accumulated rewards $R_t=\sum_{t=k}^T\gamma^{j-t} r_t$, where $\boldsymbol{s}$ is the robot's state (local observation), $\boldsymbol{a}$ is the executed action and $r_t$ is the reward obtained at timestep $t$. 

In deep Q-Learning, the objective is to maximize an action-value function, namely $Q^{\pi}(\boldsymbol{s},\boldsymbol{a})$ \cite{mnih2015human}. For an optimal policy $\pi^*$,  there is $Q^{\pi^*}(\boldsymbol{s},\boldsymbol{a})=\max_\pi\mathbb{E}[R_t |\boldsymbol{s},\boldsymbol{a},\pi]=r+{\gamma}Q^{\pi^*}(\boldsymbol{s}',\boldsymbol{a}')$, where $\boldsymbol{s}'$ and $\boldsymbol{a}'$ are the state and action at the next timestep. 
% Once $Q^{\pi_\theta^*}$ is known, the optimal policy is the one that executes the action. $\boldsymbol{a}'$ with the maximal $Q$-value. 

In real applications,  the Q-value function will be parameterized by $\boldsymbol{\theta}$. The following loss function will be minimized using a gradient descent method.
\begin{equation*}
    L_j (\theta)=\frac{1}{2}\big(y_j-Q_j(\boldsymbol{s},\boldsymbol{a};\theta)\big)^2
\end{equation*}
where $y_j=\mathbb{E}\big[r+\gamma\max_{\boldsymbol{a}'}Q_j(\boldsymbol{s}',\boldsymbol{a}')|\boldsymbol{s},\boldsymbol{a}\big]$.

\subsection{Artificial Potential Field}
The multi-robot APF method guides robots to a goal position via the combination of attractive, repulsive, and inter-individual forces \cite{koren1991potential}. In this paper, the attractive force $F_{a,i}$ is a unit vector pointing to the target. The mathematical form of the repulsive force is  
% \begin{equation}
%     \label{eq:repulse}
%     \boldsymbol{F}_{r,i}=\begin{cases}
%     \eta (\frac{1}{\|\boldsymbol{x}_{o,i}-\boldsymbol{x}_i)\|}-\frac{1}{\rho_0})\frac{\boldsymbol{x}_i-\boldsymbol{x}_{o,i}}{\|\boldsymbol{x}_{o,i}-\boldsymbol{x}_i\|^3}, & \text{if } \|\boldsymbol{x}_{o,i}-\boldsymbol{x}_i\|\leq\rho_0 \\
%     \boldsymbol{0}, & \text{if } \|\boldsymbol{x}_{o,i}-\boldsymbol{x}_i\|>\rho_0
%     \end{cases}
% \end{equation}
\begin{equation*}
    \label{eq:repulse}
    \boldsymbol{F}_{r,i}=\begin{cases}
    \eta \frac{\rho_0-\|\boldsymbol{x}_{o,i}-\boldsymbol{x}_i\|}{\|\boldsymbol{x}_{o,i}-\boldsymbol{x}_i\|^3\rho_0}\frac{\boldsymbol{x}_i-\boldsymbol{x}_{o,i}}{\|\boldsymbol{x}_{o,i}-\boldsymbol{x}_i\|}, & \text{if } \|\boldsymbol{x}_{o,i}-\boldsymbol{x}_i\|\leq\rho_0 \\
    \boldsymbol{0}, & \text{if } \|\boldsymbol{x}_{o,i}-\boldsymbol{x}_i\|>\rho_0
    \end{cases}
\end{equation*}
where $\rho_0$ is the influence range of obstacles and $\eta$ is a positive scaling factor \cite{khatib1986real}.
The inter-individual force is defined as 
\begin{equation}
    \label{eq:individual}
    \boldsymbol{F}_{in,i}=\sum_{\substack{j\neq i \\ j\in P(i)}}\Bigg(0.5-\frac{\lambda}{\|\boldsymbol{x}_j-\boldsymbol{x}_i\|}\Bigg)\frac{\boldsymbol{x}_j-\boldsymbol{x}_i}{\|\boldsymbol{x}_j-\boldsymbol{x}_i\|}
\end{equation}
where $P(i)$ is the set of teammates observed by the $i$-th robot. 

The total virtual force is, therefore,
\begin{equation}
    \boldsymbol{F}_{i}=\boldsymbol{F}_{a,i}+\boldsymbol{F}_{r,i}+\boldsymbol{F}_{in,i}
\end{equation}

\subsection{Wall Following}\label{subsec:wall following}
\begin{figure}
    \centering
    \includegraphics[width=0.45\textwidth]{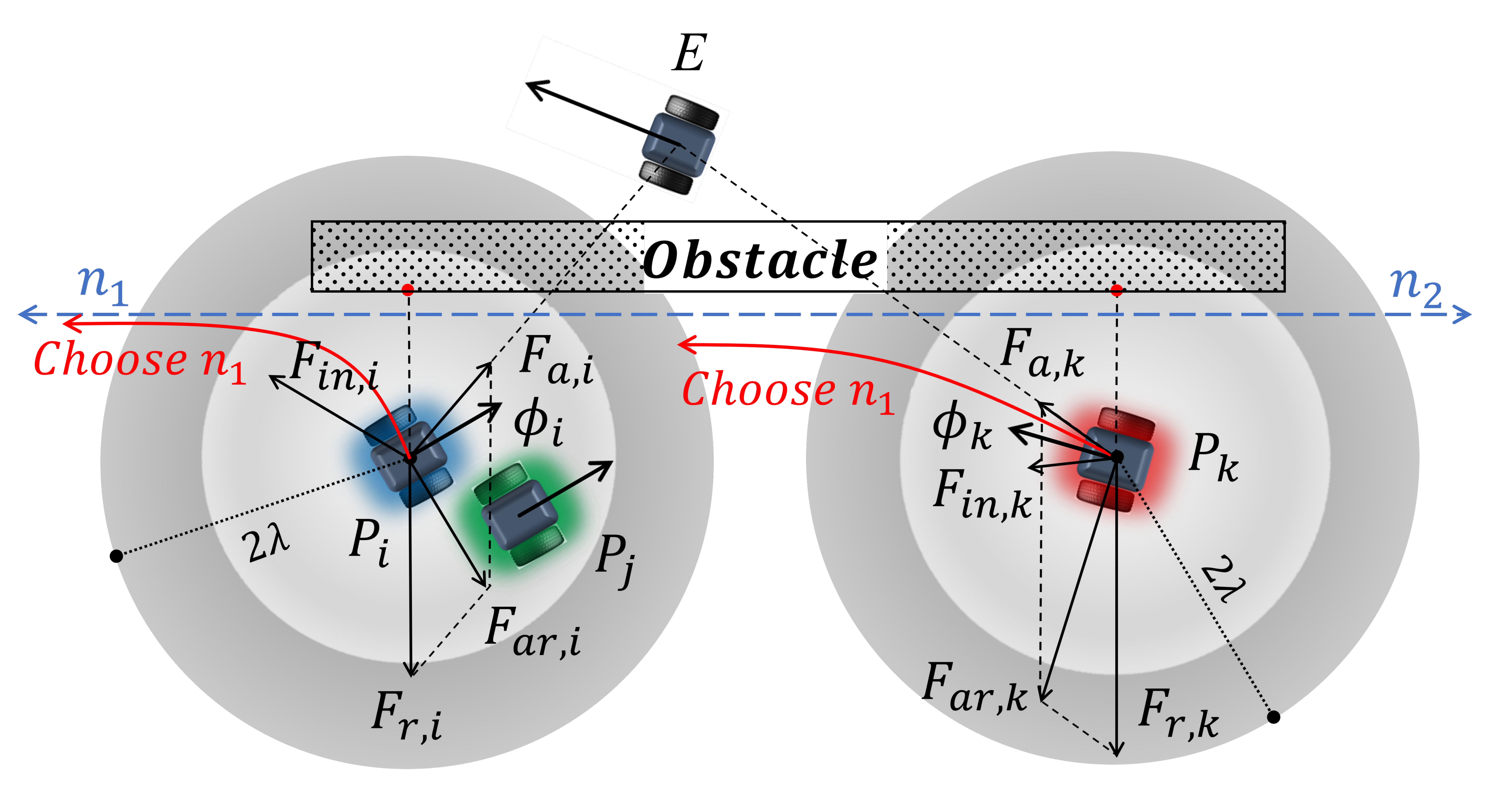}
    \caption{The wall following rules. When the angle between $\boldsymbol{F}_{ar,i}$ and $\boldsymbol{F}_{a,i}$ exceeds $90^{\circ}$,  $P_i$ moves by the wall following rules. Due to the repulsive inter-individual force from $P_j$, $P_i$ chooses to turn sharply for collision avoidance. On the contrary, $P_k$ moves in the direction of $\boldsymbol{n}_1$ for a smooth turn.}
    \label{fig:wall following}
\end{figure}

Conventional APF methods are notorious for their local minima issue \cite{ge2000new}.  In this paper, the local minima issue is mitigated significantly by using wall following rules \cite{borenstein1989real,yun1997wall}. 
% is introduced in this paper 
% The APF method has some inherent limitations \cite{ge2000new}. For example, when an obstacle is between the robot and the target, the attractive force and the repulsive force exerted on the robot are collinear and opposite, which means the robot can not bypass the obstacle and will be trapped when the total force equals 0. To avoid these trap situations due to local minima, we adapt the wall following rules proposed in \cite{borenstein1989real} and \cite{yun1997wall}.
As shown in Figure \ref{fig:wall following}, the resultant force of the attractive and repulsive forces is denoted as $\boldsymbol{F}_{ar,i}$. Before each move, the robot checks whether the angle between $\boldsymbol{F}_{ar,i}$ and $\boldsymbol{F}_{a,i}$ exceeds $90^{\circ}$. If so, the robot will move in the direction of either $\boldsymbol{n}_1$ or $\boldsymbol{n}_2$, which are vectors perpendicular to $\boldsymbol{F}_{r,i}$. The choice between $\boldsymbol{n}_1$ and $\boldsymbol{n}_2$ depends on the current heading $\phi_i$ and the inter-individual force $\boldsymbol{F}_{in,i}$. If $\boldsymbol{F}_{in,i}$ exceeds a predefined threshold $B$, the robot will select the vector which forms a smaller angle with $\boldsymbol{F}_{in,i}$, otherwise the vector which forms a smaller angle with $\phi_i$ will be chosen.

The APF methods might also experience the local minima problem when several robots are all in the proximity of the target. For example, when a robot $j$ has reached the target, it may block the way of another robot $i$ to the target. Analogous to the ``robot--obstacle--target'' situation in Figure \ref{fig:wall following}, the robot $i$ can not bypass the robot $j$ in the ``robot $i$--robot $j$--target'' situation either. Hence, a virtual obstacle is introduced in the position of any pursuer that has reached the target, which helps following robots go around teammates by the wall following rules.

%%%%%%%%%%%%%%%%%%%%%

\section{Adaptive Cooperative Pursuit Algorithm}\label{sec:acp}

\subsection{Observation Embedding}
In the multi-robot setting, the local observation's dimension of a pursuer would change with respect to time because of the possible varying number of neighbours. To address this issue,  we follow the design in \cite{huttenrauch2019deep}, which employs a one-layer fully connected neural network to extract high-dimensional features of each neighbour teammate. After that, the mean of all teammates’ features is concatenated with the local environment observations $\boldsymbol{o}_{loc,i}$ including $d_{o,i},\phi_{o,i},d_{e,i},\phi_{e,i}$. Since the resultant vector is fixed-length, we can input it directly into the policy network.

% The pursuer's local observation is the input of the policy network. Since
% the neural network policy assumes a fixed input dimensionality, it can not tackle the variable number of the teammates observed. Therefore,

\begin{figure*}
    \centering
    \includegraphics[width=0.95\textwidth]{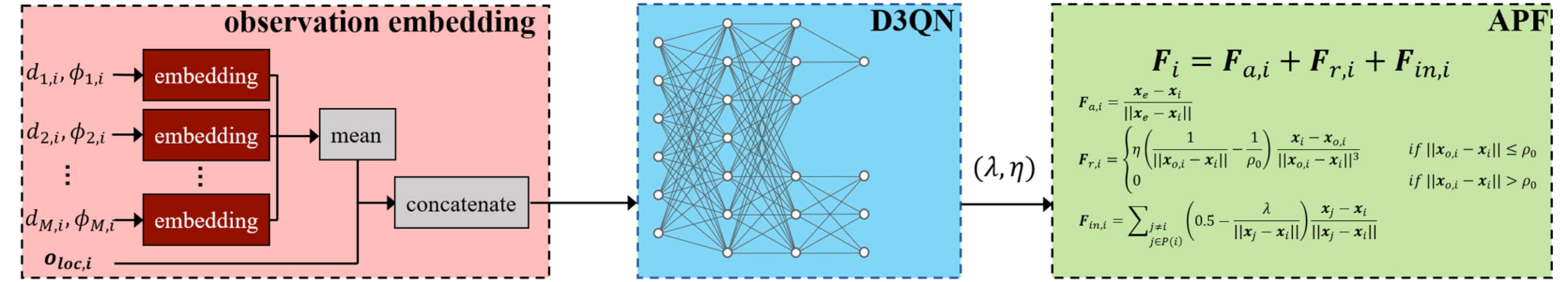}
    \caption{The deep neural network' structure of DACOOP. Brown blocks: the observation embedding which is trained following the idea in \cite{huttenrauch2019deep}. In this paper, the observation embedding is a one-layer fully connected network with 128 units. The first layer of D3QN network has the same number of units as the observation embedding, after which two streams of fully connected networks are employed. The first stream has 64 units, then outputs Q-value for each valid action, while the second stream has the same size with a state-value output. All the activation functions are ReLU nonlinearities.}
    \label{fig:D3QN network}
\end{figure*}

\begin{algorithm}
    % \caption{DACOOP}\label{alg:D3QN}
    \caption{DACOOP}\label{alg:D3QN}
    \renewcommand{\algorithmicrequire}{\textbf{Input:}}
    \renewcommand{\algorithmicensure}{\textbf{Output:}}
    \begin{algorithmic}[1]
		\Require Maximal episode length $T$, number of pursuers $N$, threshold $B$, policy update frequency $T_u$, target network update frequency $C$, replay memory $\mathcal{D}$, action-value network $Q$ with random weights $\theta$, target network $Q^-$ with weights $\theta ^-=\theta$
		\Ensure Optimal policy $Q$
        \For{$episode=1,2,...,$}
            \State Initialize pursuers' observations $\boldsymbol{o}_1,\boldsymbol{o}_2,...,\boldsymbol{o}_N$
            \For{$t=1,2,..,T$}
                \For{$i=1,2,..,N$}
                    \If{Any teammate has captured the evader}
                        \State Treat it as an obstacle
                    \EndIf
                    \State Select $(\eta_i,\lambda_i)$ by $\epsilon$-greedy \cite{mnih2015human}
                    \State Calculate attractive force $\boldsymbol{F}_{a,i}$, $\boldsymbol{F}_{r,i}$, $\boldsymbol{F}_{in,i}$.
                        \If {$\angle(\boldsymbol{F}_{ar,i},\boldsymbol{F}_{a,i})<90^{\circ}$}
                            \State $\boldsymbol{F}_{i}=\boldsymbol{F}_{a,i}+\boldsymbol{F}_{r,i}+\boldsymbol{F}_{in,i}$
                        \Else
                            \State Calculate $\boldsymbol{n}_1$, $\boldsymbol{n}_2$ orthogonal to $\boldsymbol{F}_{r,i}$
                            \If {$\Vert\boldsymbol{F}_{in,i}\|<B$}
                                 \State Choose  $\boldsymbol{F}_{i}=\boldsymbol{n}_k$ by $\phi_i$
                            \Else 
                                \State Choose $\boldsymbol{F}_{i}=\boldsymbol{n}_k$ by $\boldsymbol{F}_{in,i}$
                            \EndIf
                        \EndIf
                    
                    \State Move along the direction of $\boldsymbol{F}_i$
                    \State Observe the reward $r_i$ and new observation $\boldsymbol{o}_i'$
                    \State Store the transition $(\boldsymbol{o}_i,\boldsymbol{a}_i,r_i,\boldsymbol{o}_i')$ in $\mathcal{D}$
                \EndFor
            \EndFor
            \For{$update=1,2,...T_u$}
                \State Sample transitions from $\mathcal{D}$ as \cite{schaul2015prioritized}
                \State Calculate loss as \cite{wang2016dueling} and update the network $Q$
                \State Every $C$ steps copy $\theta$ to $\theta^-$
            \EndFor
        \EndFor
    \end{algorithmic}
\end{algorithm}

\subsection{D3QN-based Adaptive Cooperative Pursuit Algorithm}

In the traditional APF method, several parameters are so influential that they determine the behaviors of the pursuers. Particularly, the scale factor of the repulsive force $\eta$ adjusts the influence of obstacles while the parameter $\lambda$ in (\ref{eq:individual}) regulates the tightness of the multi-robot team. Note that $\lambda$ reflects pursuit strategies to some extent. A larger $\lambda$ means the pursuers repulse each other to surround the evader while a smaller $\lambda$ leads to encirclement contraction. In our design, the action space is chosen to be the parameter pair $(\eta,\lambda)$, so one has $\pi_\theta(\eta,\lambda|\boldsymbol{s})$. The APF is thereafter employed to compute the resultant force that determines the reference heading for pursuers.

The action space is discretized into $H$ parameter pairs empirically. Typically, $\eta$ ranges from $0$ (totally ignoring obstacles) to a very large value (highly conservative motions for obstacle avoidance), while $\lambda$ ranges from a small positive value (moving as a tight group) to some large one (repulsing teammates in the distance).

\subsection{Reward Design}
\begin{figure}
    \centering
    \subfigure[]{
    \includegraphics[width=0.14\textwidth]{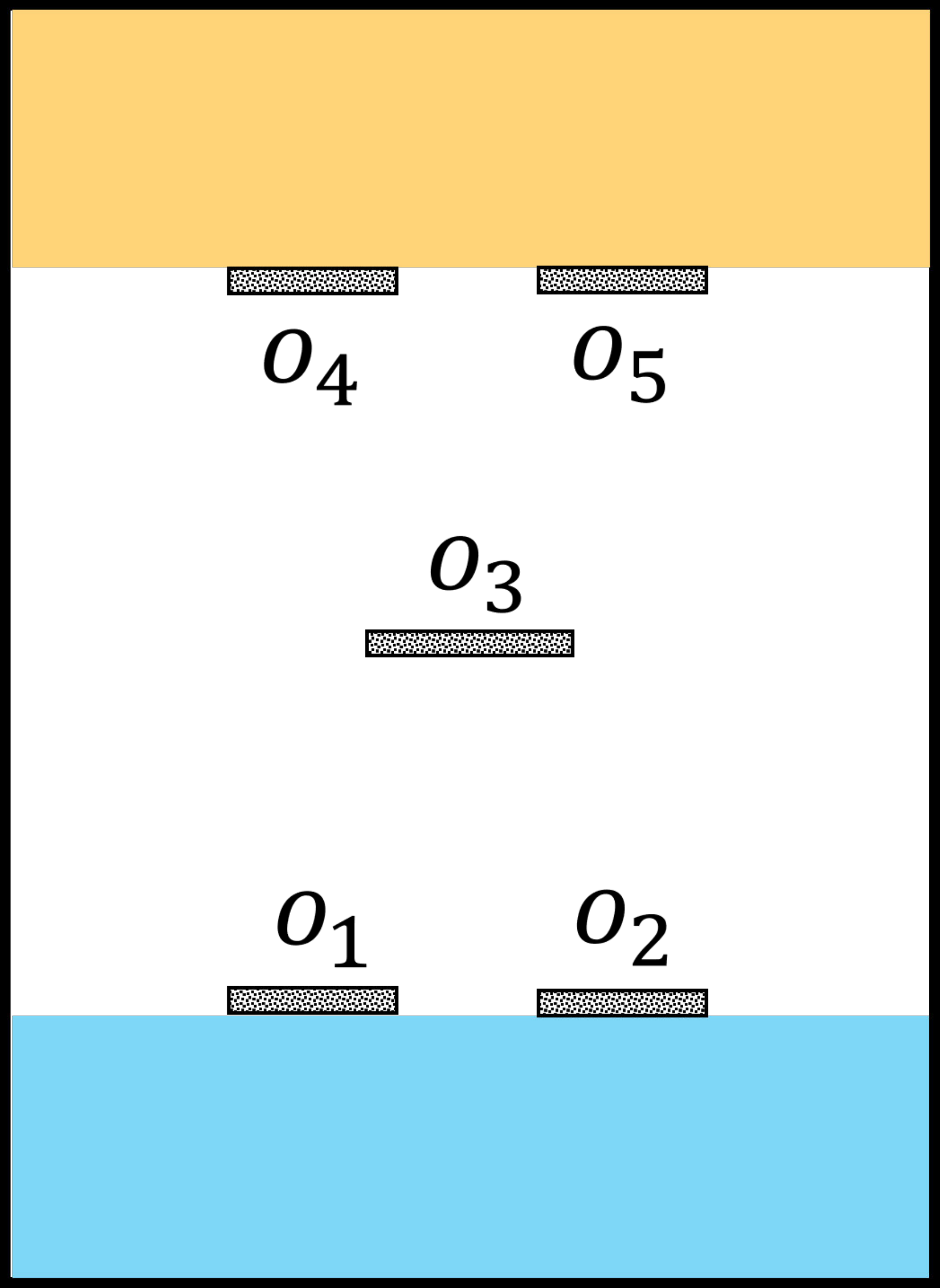}}
    \subfigure[]{
    \includegraphics[width=0.14\textwidth]{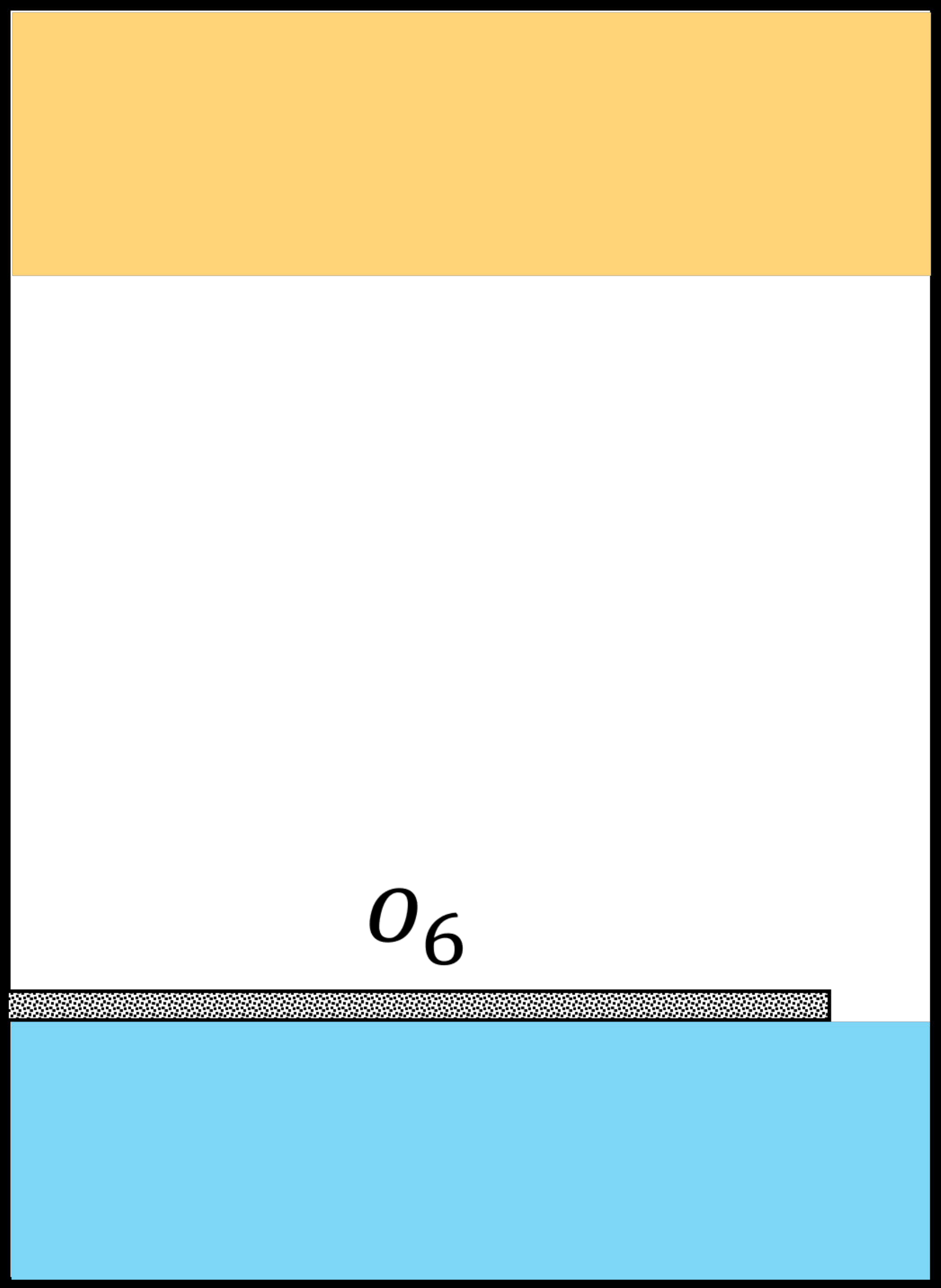}}
    \caption{The cooperative pursuit arenas. We train the cooperative pursuit policy in (a) while verifying the generalization ability in (b). The pursuers are randomly initialized in the blue area while the evader in the yellow area. $o_1$, $o_2$, $o_3$, $o_4$, $o_5$, and $o_6$ are obstacles.}
    \label{fig:pursuit arena}
\end{figure}

In our proposed algorithm, each pursuer obtains its own reward at each timestep, which is defined as
\begin{equation*}
    r=r_{main}+r_{time}+r_{tm}+r_{o}+r_{pot}
\end{equation*}
where $r_{main}$ is $20$ when the pursuer captures the evader and $0$ otherwise.
The reward $r_{time}$ is employed to encourage pursuers to move smoothly, so it gives a penalty of $-5$ when the difference of headings between adjacent timesteps exceeds $45^\circ$. When the pursuer collides with other teammates, $r_{tm}$ gives a penalty of $-20$. The pursuer will be punished by $r_{o}$ if it is too close to obstacles, which is defined as 
\begin{equation*}
    r_{o}^t=\begin{cases}-20, &\text{if }d_{o}<d_p \\
    -2, &\text{if }d_p\leq d_{o}<1.5d_p\\
    0, &\text{otherwise}\end{cases}
\end{equation*}
The  function $r_{pot}$ is a potential-based reward with 
\begin{equation*}
    r_{pot}=\gamma\Phi(\boldsymbol{s})-\Phi(\boldsymbol{s}'')
\end{equation*}
where $\boldsymbol{s}$ is the robot's state \cite{ng1999policy}, and 
\begin{equation*}
    \Phi(\boldsymbol{s})=\begin{cases}15, &\text{if }d_{e}<0.4 \text{ m} \\
    10, &\text{if }0.4 \text{ m}\leq d_{e}<0.6 \text{ m}\\
    5, &\text{if }0.6 \text{ m}\leq d_{e}<0.8 \text{ m}\\
    0, &\text{otherwise}\end{cases}
\end{equation*}
Note that the pursuit arena is $3.6 $ m $\times$ $5$ m (c.f. Figure \ref{fig:pursuit arena}), so the pursuer is only rewarded by $r_{pot,i}$ when it is close to the evader. Hence, the flanking behavior will not be punished, even if it makes the pursuer temporarily away from the evader.

\subsection{Training Setup}
Our algorithm adopts the \emph{centralized training decentralized execution} paradigm with parameter sharing, which implies a shared policy is trained with experiences collected by all pursuers. The robust reinforcement learningÒ algorithm, D3QN \cite{wang2016dueling}, is extended to the multi-robot scenario in an independent RL manner \cite{matignon2012independent}, which treats the teammates and evader as a part of the external environment. Consequently, the whole pursuit policy is composed of the observation embedding, D3QN network and APF. The proposed algorithm is named DACOOP (D3qn-based Adaptive COOperative Pursuit algorithm) with the network architecture depicted in Fig. \ref{fig:D3QN network}. The whole training procedure is summarized in Algorithm \ref{alg:D3QN}.

%%%%%%%%%%%%%%%%%%%%%%%%%%%%%

%%%%%%%%%%%%%%%%%%%%%%%%%%%%%%
\section{Results And Discussion}\label{sec:result}
In this section, simulations and experiments are performed to demonstrate the efficiency of the proposed DACOOP algorithm. In the simulations, the advantages of DACOOP are verified by comparing with an improved APF method and a vanilla RL algorithm. In the experiments, the learned policies by DACOOP are implemented on ground robots to illustrate their efficiency in real-life applications. In both simulations and experiments, the escaping policy of the evader is a modified version of the one in \cite{janosov2017group}. 

% The learned policy is employed in differential wheeled robots using direct sim-to-real transfer.

% The escape policy is fixed and can be refered in {\color{red} ?}.

\subsection{Numerical Simulations}\label{subsec:simulation}
The pursuit arena is shown in Fig. \ref{fig:pursuit arena}(a). The pursuers’ speed is $0.3$ $m/s$ while the faster evader’s speed is $0.4$ $m/s$ .  The action space of DACOOP has $24$ parameter pairs combined with $\eta=0$, $1.5 \times 10^8$, $3\times10^8$ and $\lambda=30$, $100$, $250$, $500$, $750$, $1000$,  $2000$, $3000$. Each episode is run maximally $1000$ steps. The Q-value neural network is trained using Adam optimizer until the success rate converges (approximately $7000$ episodes) \cite{kingma2014adam}. We update the network $1000$ times after each episode, i.e. $T_u$ in Algorithm \ref{alg:D3QN} is 1000. The learning rate is $3\times10^{-4}$. The discount factor $\gamma$ is $0.99$. The $\epsilon$-greedy exploration is employed with $\epsilon$ linearly decayed from $1$ to $0.01$ in $4000$ episodes. The DACOOP algorithm is compared with the following methods.
\begin{itemize}
    \item D3QN: selecting a reference heading from 24 uniformly distributed directions. An extra reward $r_{app,i}=\frac{d_{e,i}''-d_{e,i}}{200}$ is introduced with $d_{e,i}''$ denoting the distance to the evader at the last timestep. Besides the action space and reward function, all settings are the same as DACOOP;
    \item Modified APF: reducing $\eta$ and $\lambda$ from their initial values to $0$ with the decrease of the distance to the evader. The best initial values are chosen from the candidate parameter pairs of DACOOP.
\end{itemize}

%  a simple heuristics of adjusting hyperparameters, linearly decaying
%We use $Success$ denotes episodes in which the mission is complete, $Collision$ denotes episodes in which collisions occur, and $Time$-$out$ denotes those episodes in which the mission is not completed but there is no collision.

\subsubsection{Sample efficiency}
\begin{figure}
    \centering
    \includegraphics[width=0.38\textwidth]{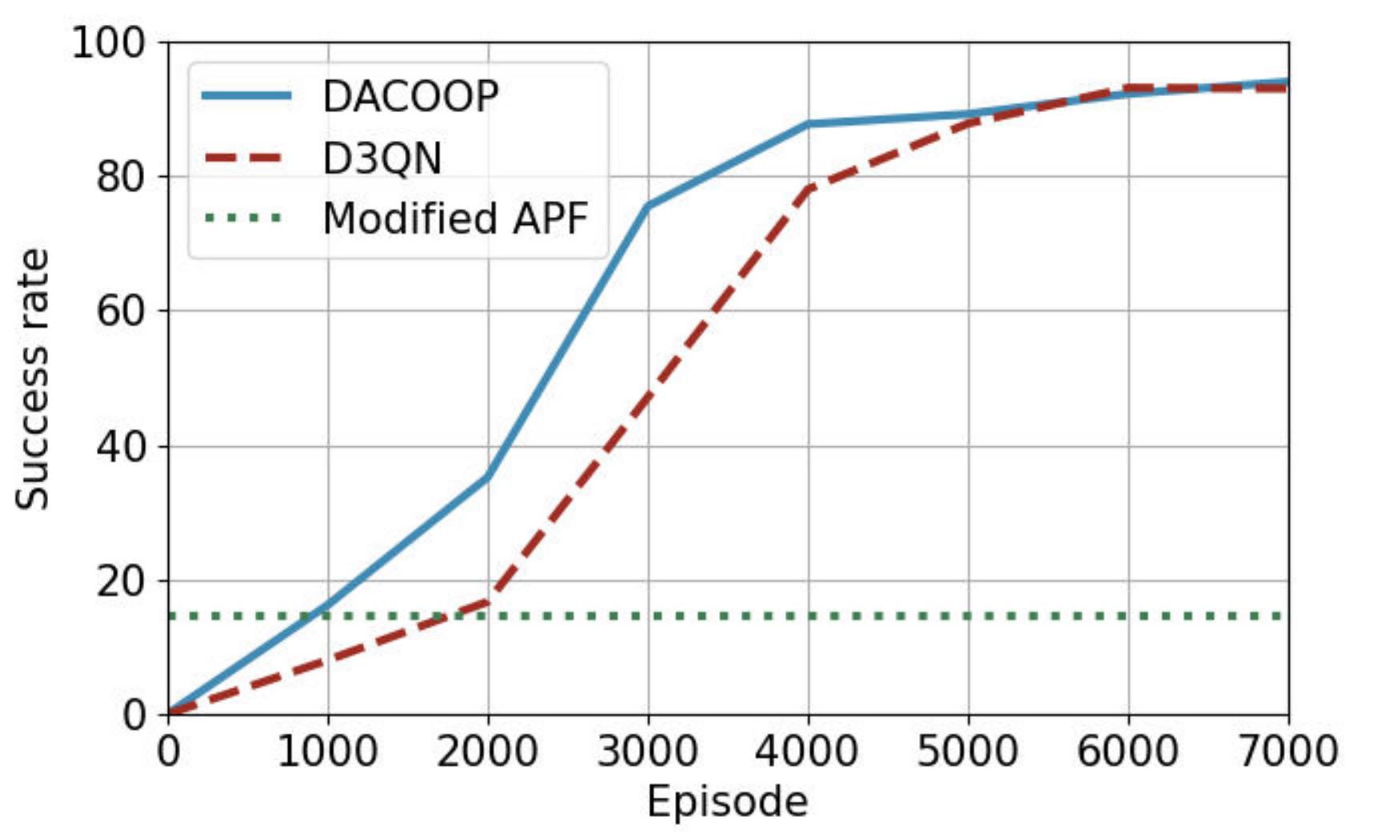}
    \caption{The success rates of DACOOP, D3QN and the modified APF for every 1000 episodes at training.}
    \label{fig:success}
\end{figure}

For DACOOP and D3QN, the policies are recorded every 1000 episodes at training. Fig. \ref{fig:success} shows the comparison results of the DACOOP, D3QN and modified APF over 1000 validation episodes. The final success rate of DACOOP is 94\%, which is significantly better than the modified APF (14.6\%). It verifies the significance of changing parameters for APF methods in the cooperative pursuit scenario. But it is challenging to  manually design an optimal rule that adjusts the parameters according to local observations. Both DACOOP and D3QN result in similar success rates at the end of training (93\% for D3QN). However, the proposed DACOOP is more data-efficient at learning according to Fig. \ref{fig:success}. The main reason is the attractive and repulsive forces produced by the APF layer in DACOOP help pursuers approach the evader and avoid collisions in most scenarios, so many unnecessary mistakes are excluded during exploration. Furthermore, staying away from teammates allows pursuers to cover a larger area and block more possible escape routes of an evader, which is the essence of many cooperative pursuit strategies and easily accomplished by adjusting $\lambda$. Hence, high-quality data accelerates the learning process.

% It is the essence of many cooperative pursuit strategies and easily accomplished by adjusting $\lambda$. Hence, more high-quality data accelerates the learning process. 

% is difficult, which often ends up with a sub-optimal policy as Modified APF.  reaches a comparable success rate as DACOOP at the end of training (93\%)
%  shows that D3QN has lower data efficiency
% Temporally adjusting parameters results in more efficient pursuit strategies. 

\subsubsection{Generalization}

\begin{figure}
    \centering
    \includegraphics[width=0.38\textwidth]{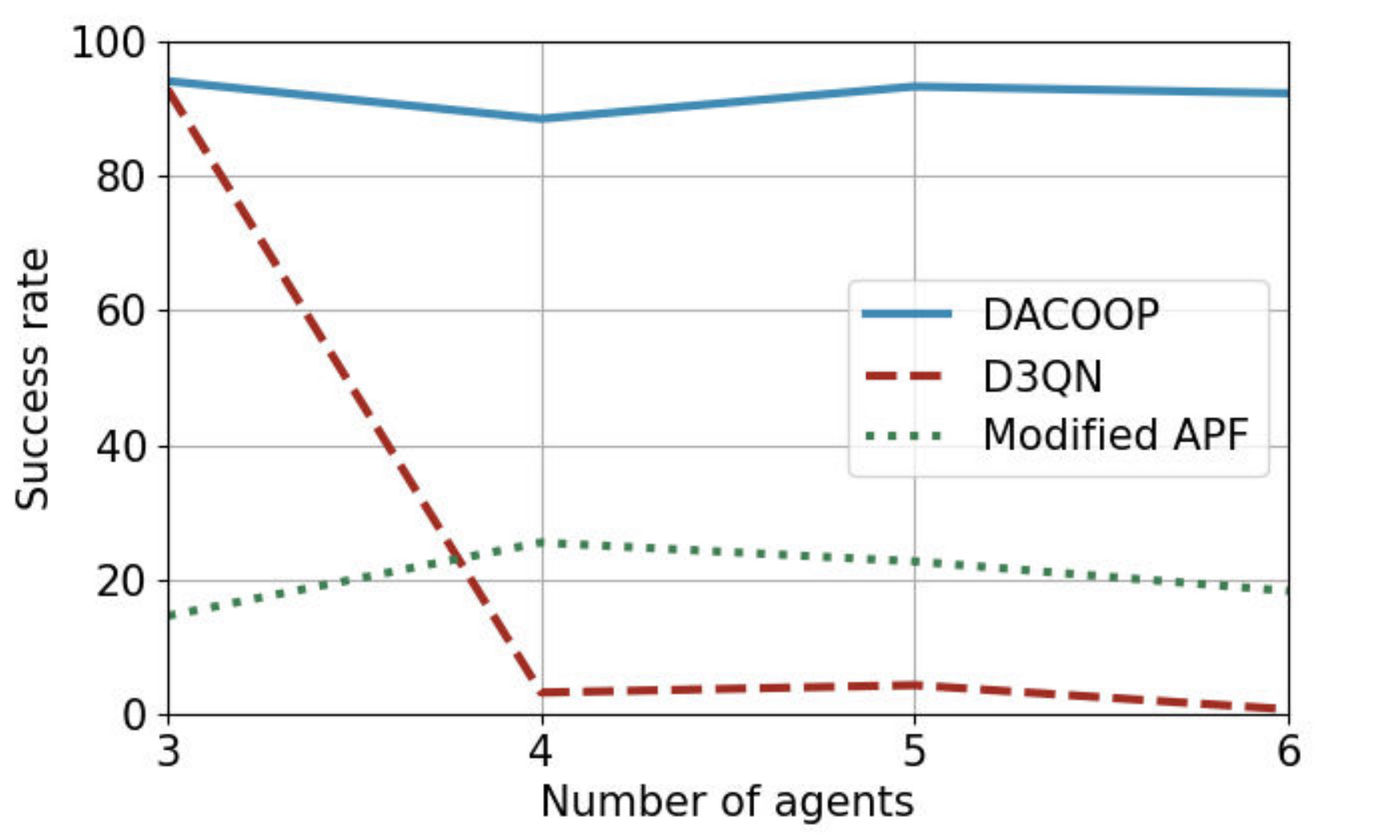}
    \caption{The success rate with different number of pursuers. All results are averaged over 1000 episodes.}
    \label{fig:generalize number}
\end{figure}

The best immediate pursuit policies trained by DACOOP and D3QN, which achieve the highest success rate in the training environment, are selected to test their generalization performance in different scenarios. For the modified APF, the initial parameter values are the same as those in Fig. \ref{fig:success}. As shown in Fig. \ref{fig:generalize number}, the policies are tested in scenarios with 3, 4, 5 and 6 pursuers, respectively. The modified APF achieves a steady performance with about 20\% success rate in all scenarios. However, the success rate of D3QN decreases dramatically with the increase of the number of pursuers. The main reason is that the empirical mean embedding becomes inaccurate when the size of the multi-robotic system changes, resulting in frequent collisions between pursuers. Although DACOOP suffers from the same issue, the pursuit policy maintains the ability of approaching the evader and avoiding collisions due to the APF layer (c.f. Fig. \ref{fig:D3QN network}). As a result, DACOOP has a better generalization performance over D3QN. 

%Fig. \ref{fig:generalize obstacle} shows the performance of different algorithms in both training and validation environments. 

\begin{figure}
    \centering
    \includegraphics[width=0.38\textwidth]{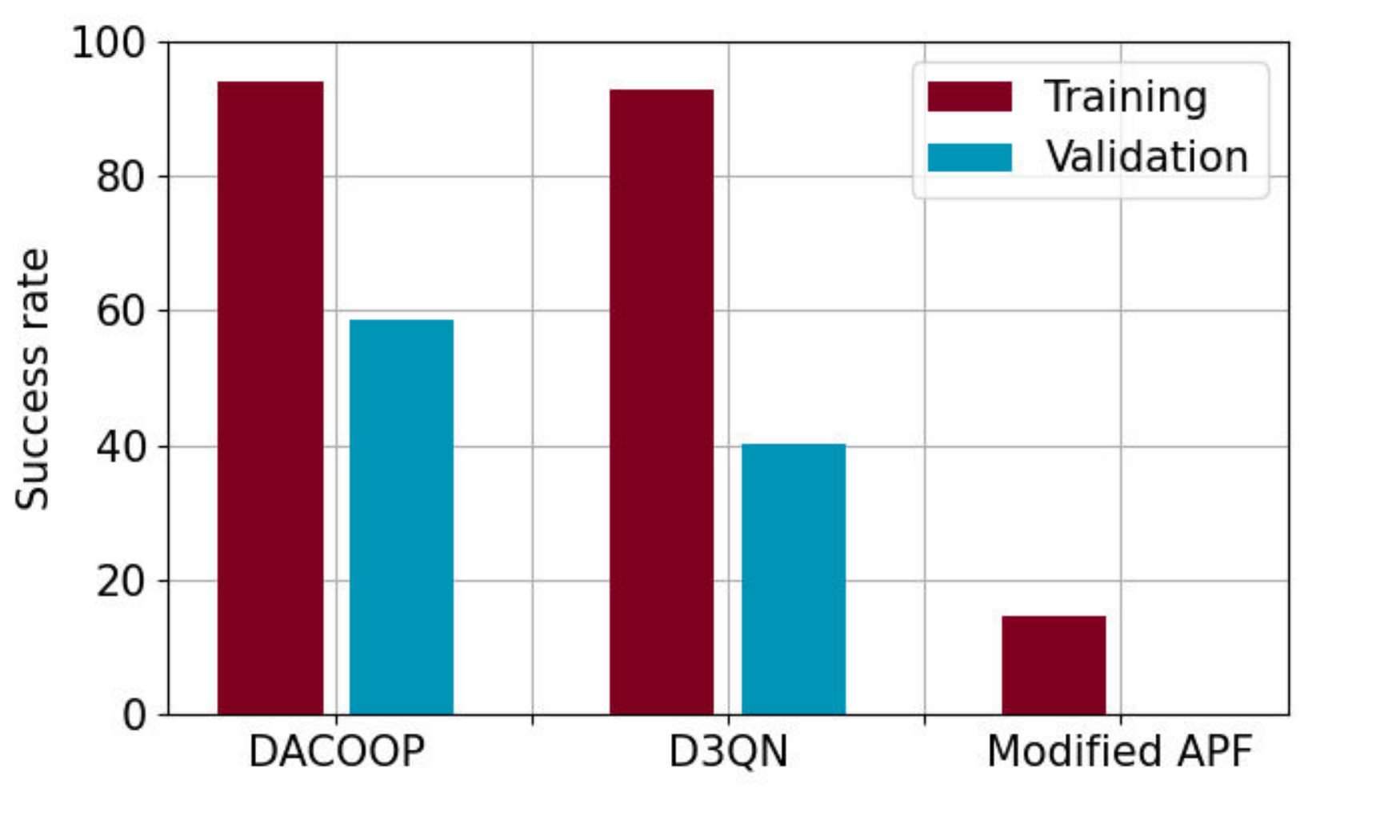}
    \caption{The success rate in both training and validation environments. The red bars denote the performance in Fig. \ref{fig:pursuit arena}(a) while the blue bars indicate the success rate in Fig. \ref{fig:pursuit arena}(b).}
    \label{fig:generalize obstacle}
\end{figure}

The learned policies are also implemented in different arenas (Fig. \ref{fig:pursuit arena}(b)). To accomplish the mission, pursuers need to pass through the narrow gap in the right of obstacle $o_6$ one by one. Note this gap is only $0.4$ m while those in the training environment are $0.9$ m and $0.5$ m (the left side of $o_1$, the right side of $o_2$ and the middle between $o_1$ and $o_2$). As shown in Fig. \ref{fig:generalize obstacle},  the success rate of the modified APF is $0$  in this validation environment. It implies inappropriate parameters hinder pursuers to pass through a narrower gap. In comparison with D3QN, the pursuit policy trained by DACOOP is less deteriorating in terms of success rate ($40$\% for D3QN and $58$\% for DACOOP). This is because the wall following rules encourage pursuers to move along obstacles until finding a way out, while D3QN pursuers are always stuck around the obstacle previously with a gap in the training environment.

\begin{figure}
    \centering
    \includegraphics[width=0.42\textwidth]{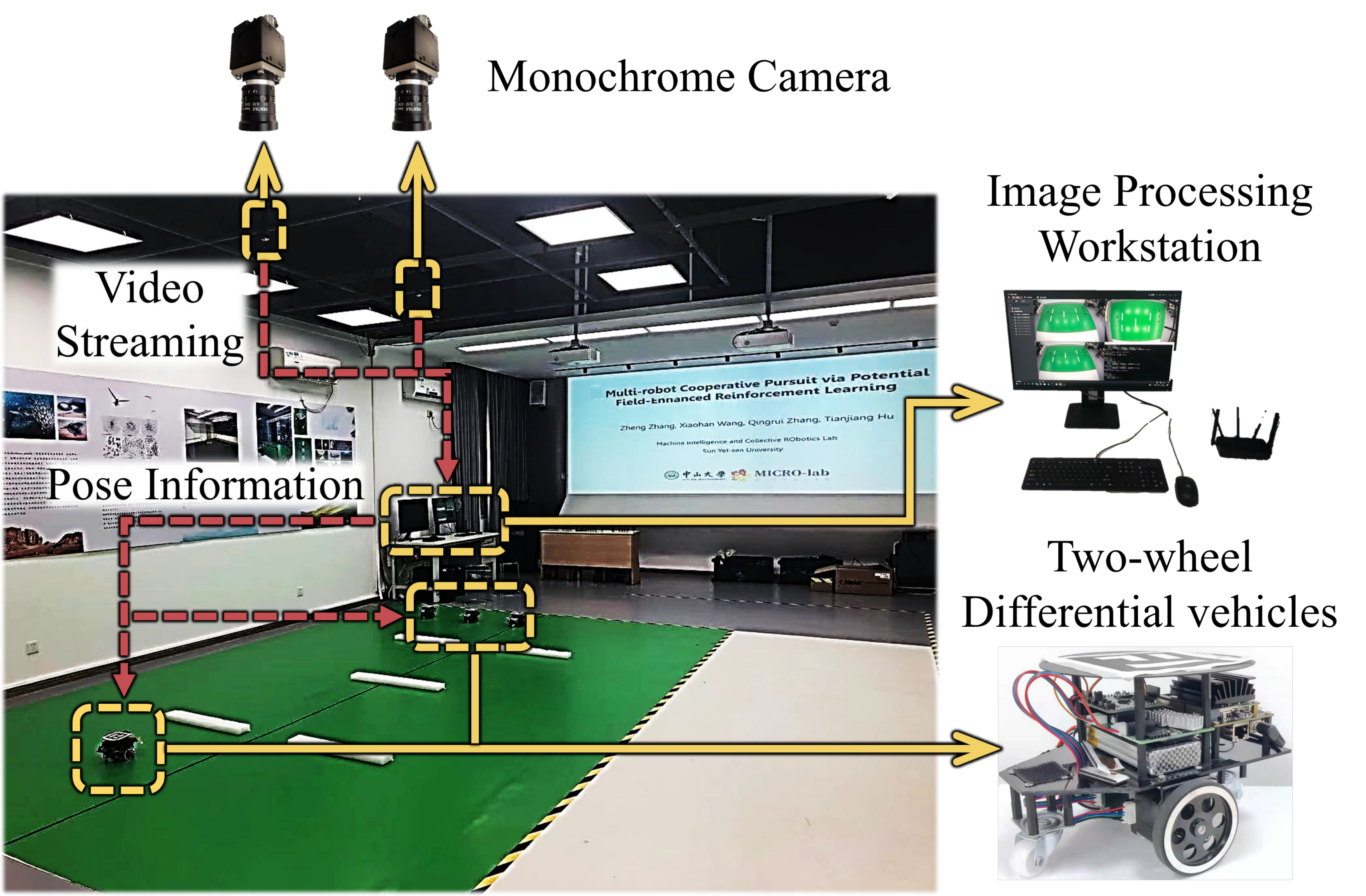}
    \caption{The multi-ground-robots experiment system.}
    \label{fig:platform}
\end{figure}

\subsection{Experiments}\label{subsec:experiment}
Our pursuit policies are applied to differential wheeled robots. Each differential wheeled robot has a high-level computation module for decision making. A low-level PID controller runs onboard to track the heading reference command. An AprilTag is attached to the top of each robot for position and heading measurement by two monochrome cameras as shown in Fig. \ref{fig:platform}. An image processing workstation receives the video streaming from the monochrome cameras and transmits the available information to each robot at $15$ Hz. The whole experiment system is shown as Fig. \ref{fig:platform}. In the rest of this section,  three sets of snapshots from different episodes are presented to demonstrate the strategies learned via DACOOP.

\begin{figure}
    \centering
    \subfigure[$t=10.16$ s]{
    \includegraphics[width=0.136\textwidth]{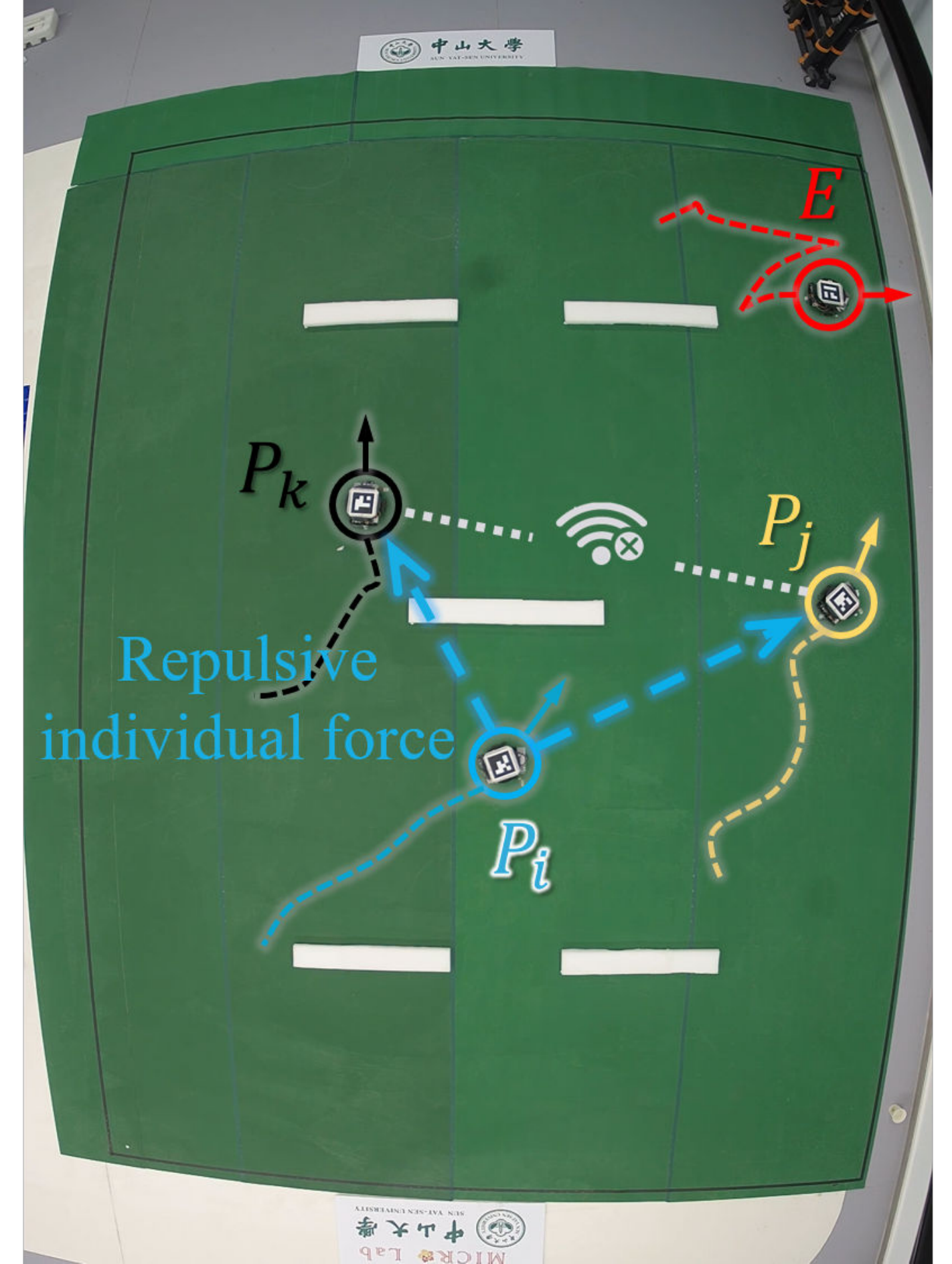}}
    \subfigure[$t=20.09$ s]{
    \includegraphics[width=0.13\textwidth]{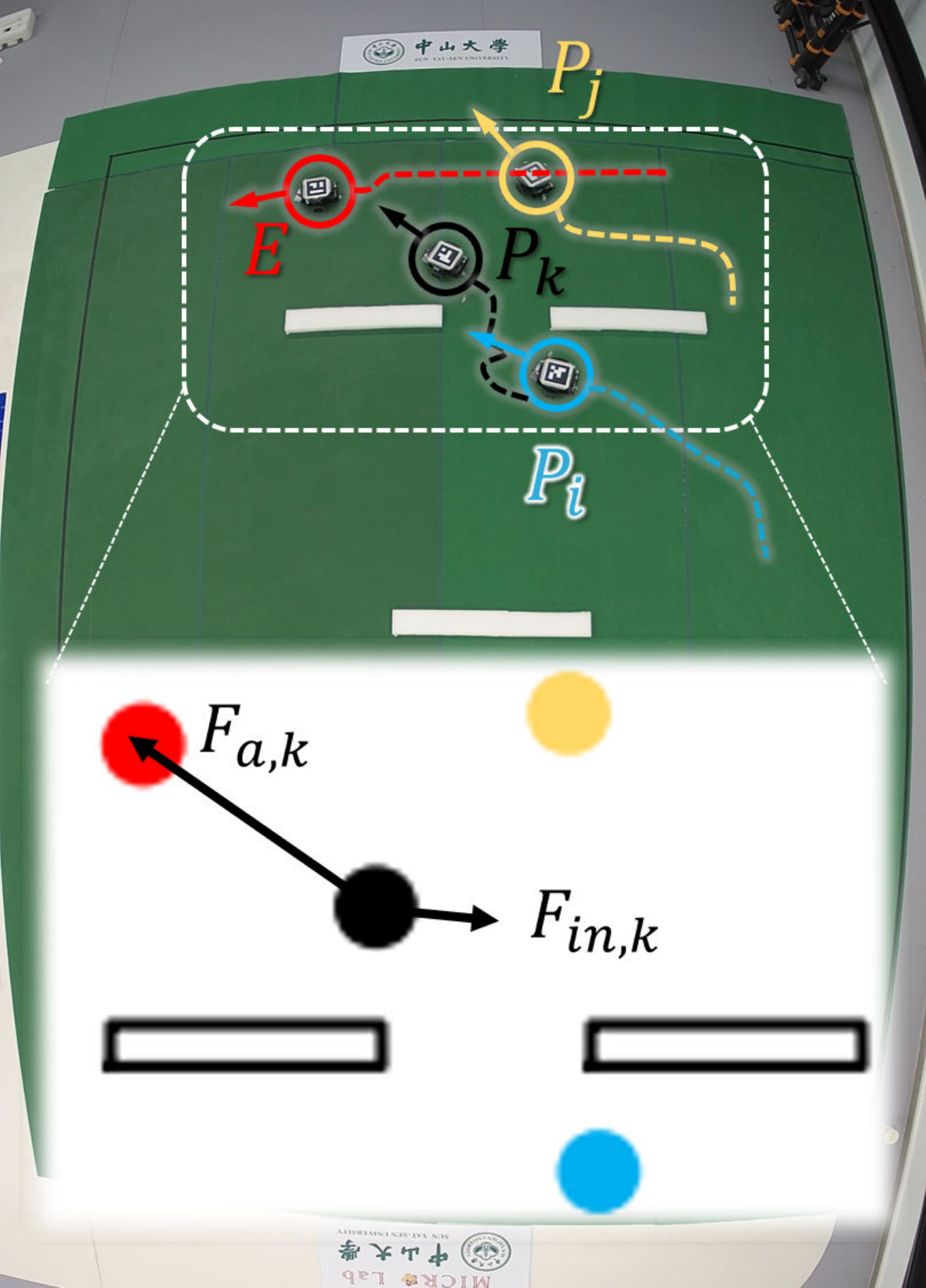}}
    \subfigure[$t=26.17$ s]{
    \includegraphics[width=0.13\textwidth]{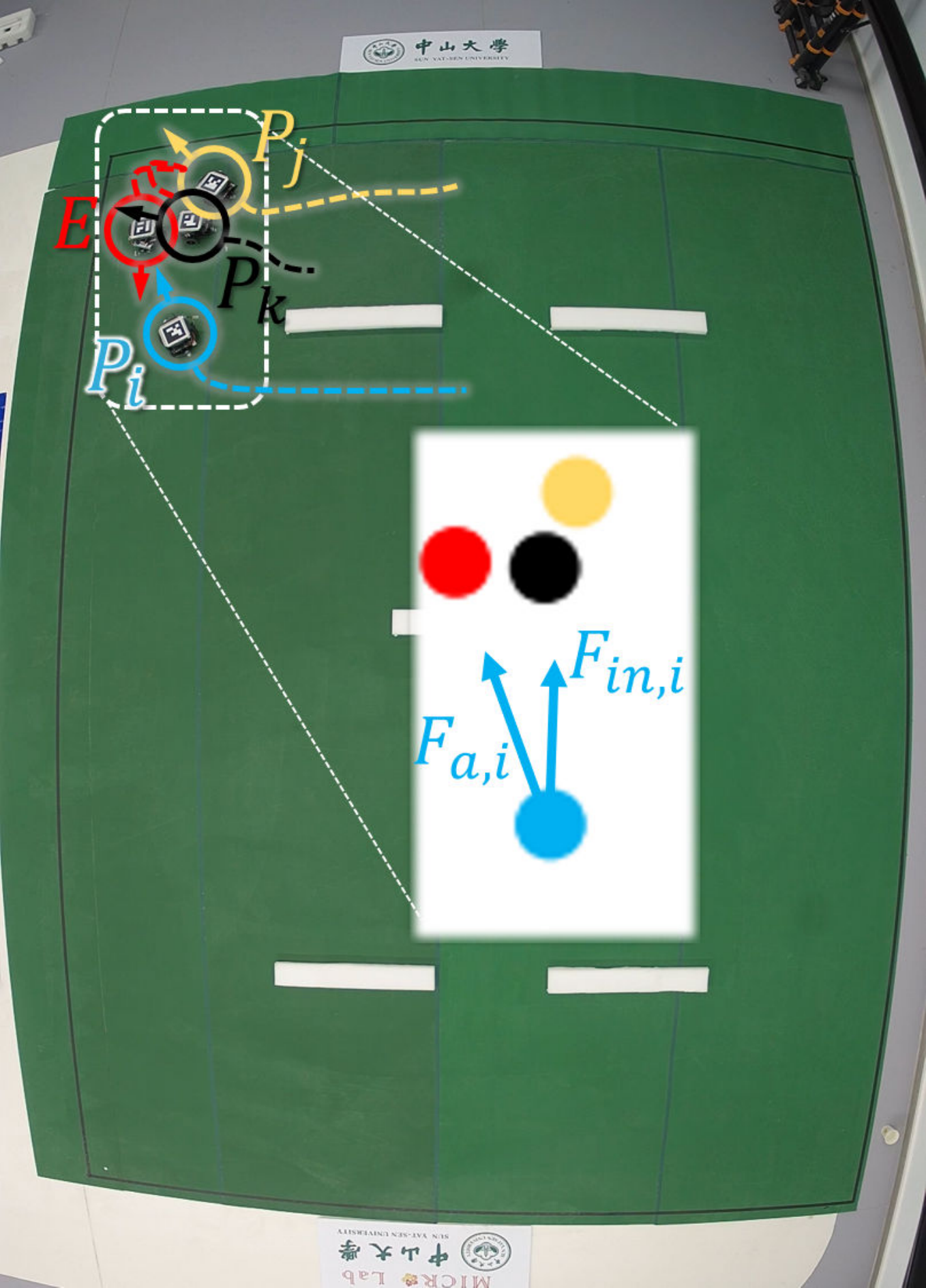}}
    \caption{The adaptive pursuit policy according to the distance from evader.}
    \label{fig:distance from evader}
\end{figure}

In the first set of snapshots, when the pursuers are far away from the evader, the pursuer $P_k$ is on the leftmost side of the team while $P_j$ is on the rightmost (Fig. \ref{fig:distance from evader}(a)). They can not detect each other as they are out of the sensing range. However, both of them can observe $P_i$ who is in the rear of the team. Therefore, a large $\lambda$ is selected to make the lateral pursuers move away from the group center, leading to the encirclement of the evader. When the pursuers are close to the evader, $P_k$ and $P_j$ approach the evader vibrantly from different paths by choosing a small $\lambda$ (Fig. \ref{fig:distance from evader}(b)). After that, the evader is captured successfully in Fig. \ref{fig:distance from evader}(c). This case shows the learned pursuit policy is adaptive to the distance from evader.  

\begin{figure}
    \centering
    \subfigure[$t=24.05$ s]{
    \includegraphics[scale=0.18]{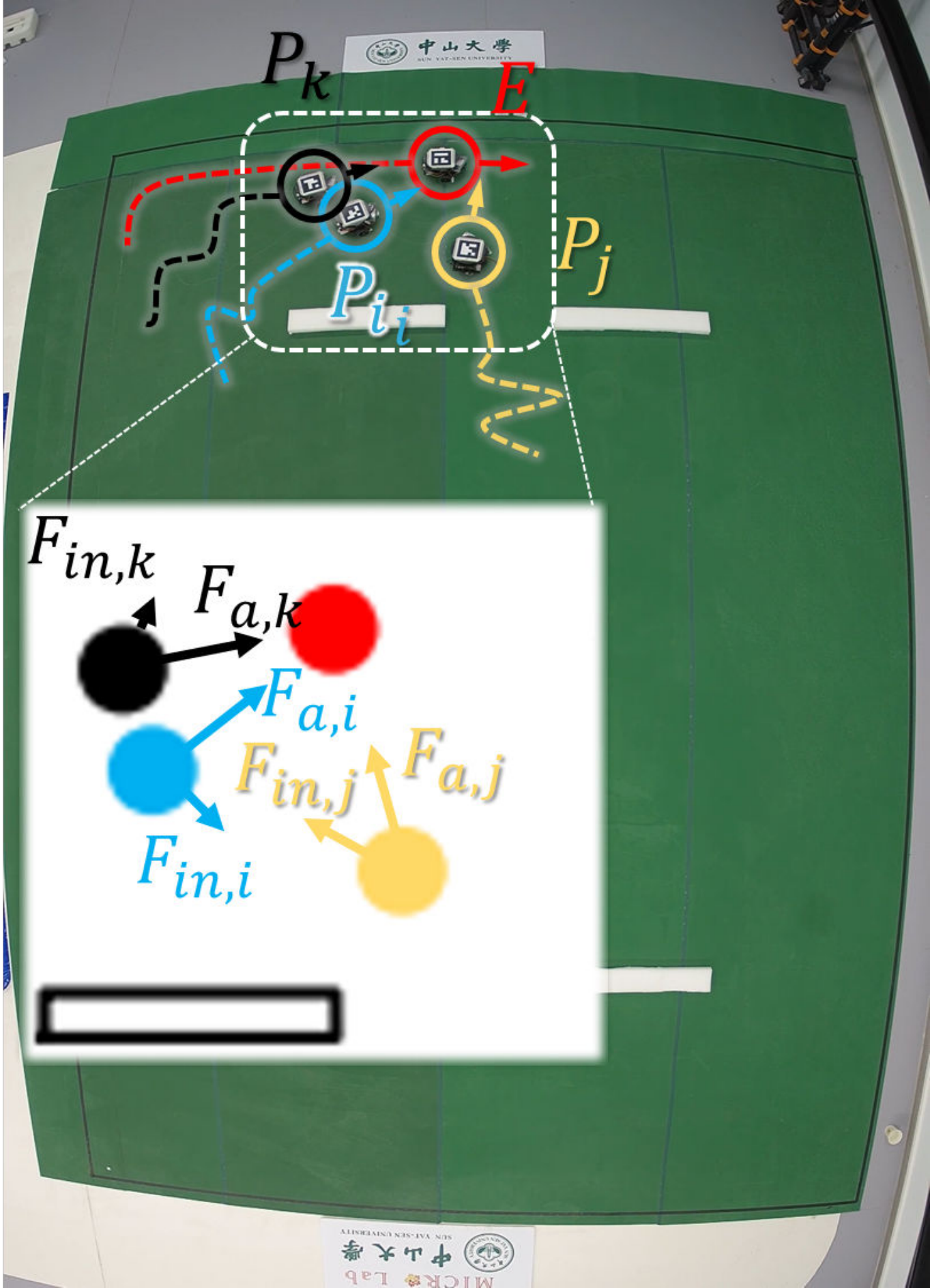}}
    \subfigure[$t=27.02$ s]{
    \includegraphics[scale=0.18]{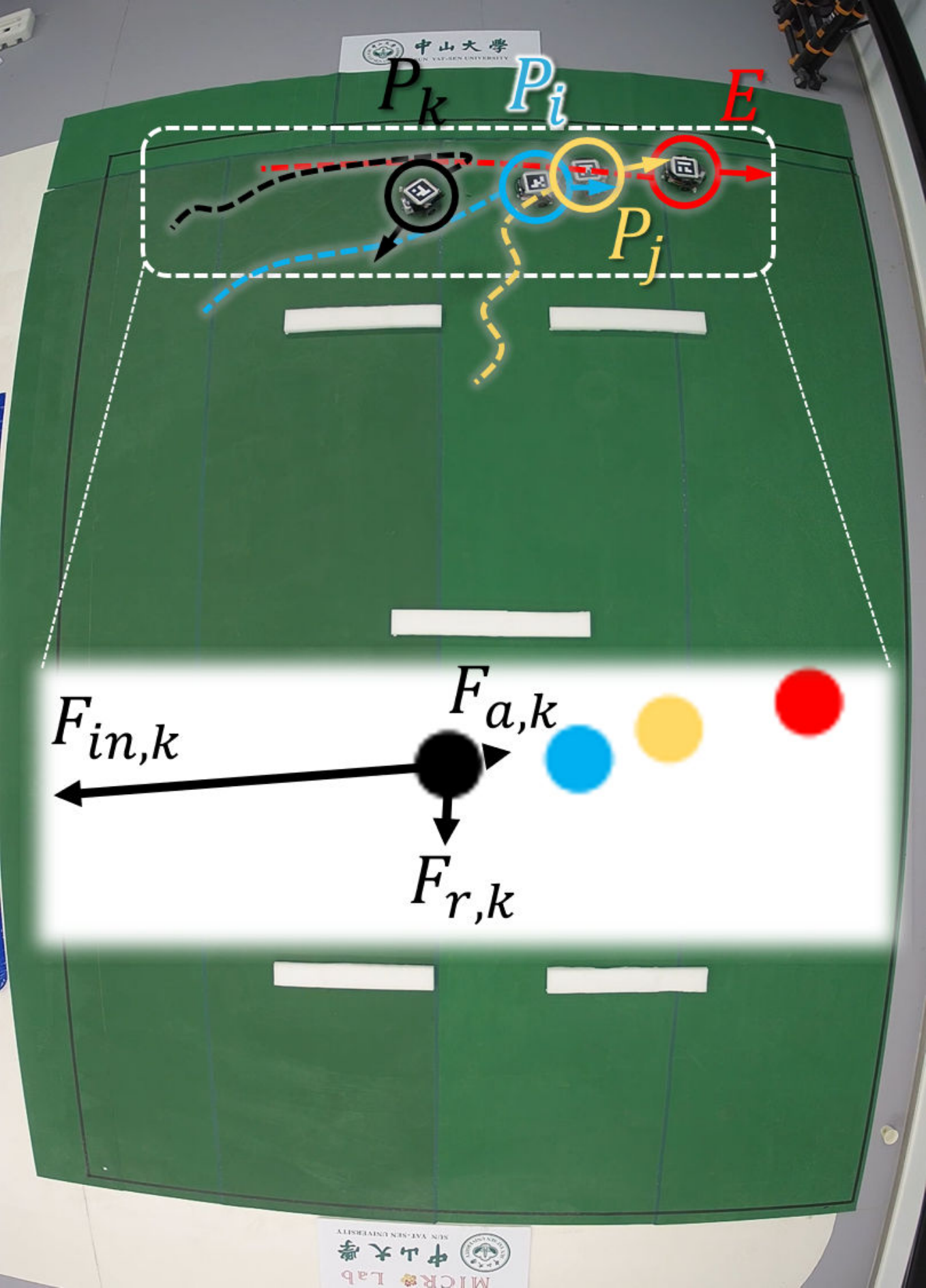}}
    \subfigure[$t=33.08$ s]{
    \includegraphics[scale=0.18]{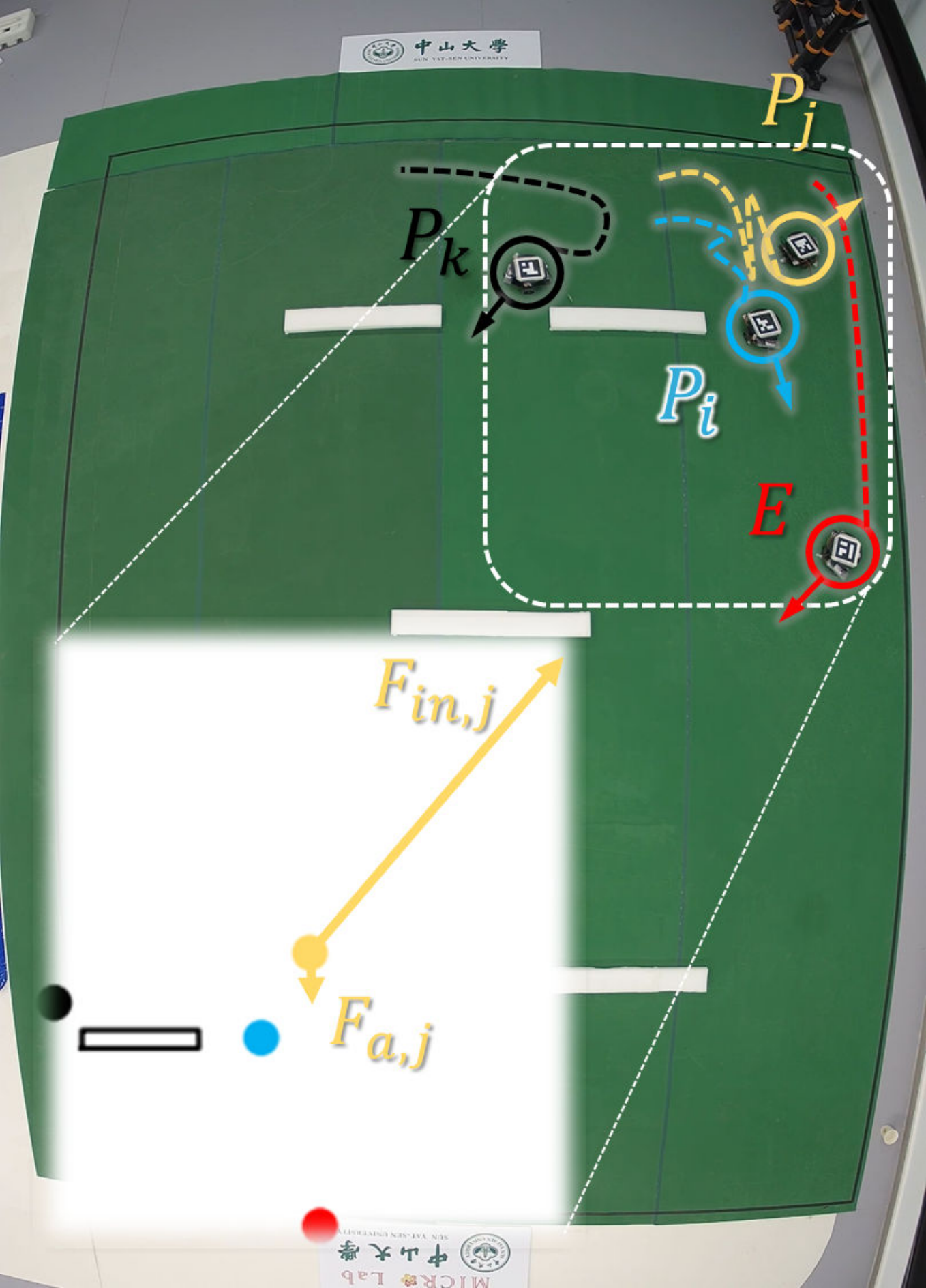}}
    \caption{The adaptive pursuit policy according to the current task. Note that the attractive force is always a unit vector and all the forces are scaled to fit the figure's size.}
    \label{fig:current task}
\end{figure}
The second set of snapshots is shown in Fig. \ref{fig:current task}. In Fig. \ref{fig:current task}(a), $P_k$ and $P_i$ chase the evader from the left, while $P_j$ flanks it from below. Since the evader is faster than pursuers, it escapes from the encirclement successfully. All pursuers are thereafter behind the evader and struggle to capture it in a short time.  Due to the short distance between pursuers, $P_k$ and $P_j$ select a large $\lambda$ to  move away from the overly close teammates for collision avoidance (Fig. \ref{fig:current task}(b) and (c)).  This case shows that the policy trained by DACOOP is adaptive to the current task. If there is a great chance to capture the evader, pursuers will take a chance to aggregate.  If the current task is to avoid collisions and wait for the next opportunity, they will repulse each other to keep a safe distance.
%Although this behavior leads to a larger distance from the evader, it eliminates the risk of collision with teammates.

\begin{figure}
    \centering
    \subfigure[$t=20.13$ s]{
    \includegraphics[scale=0.18]{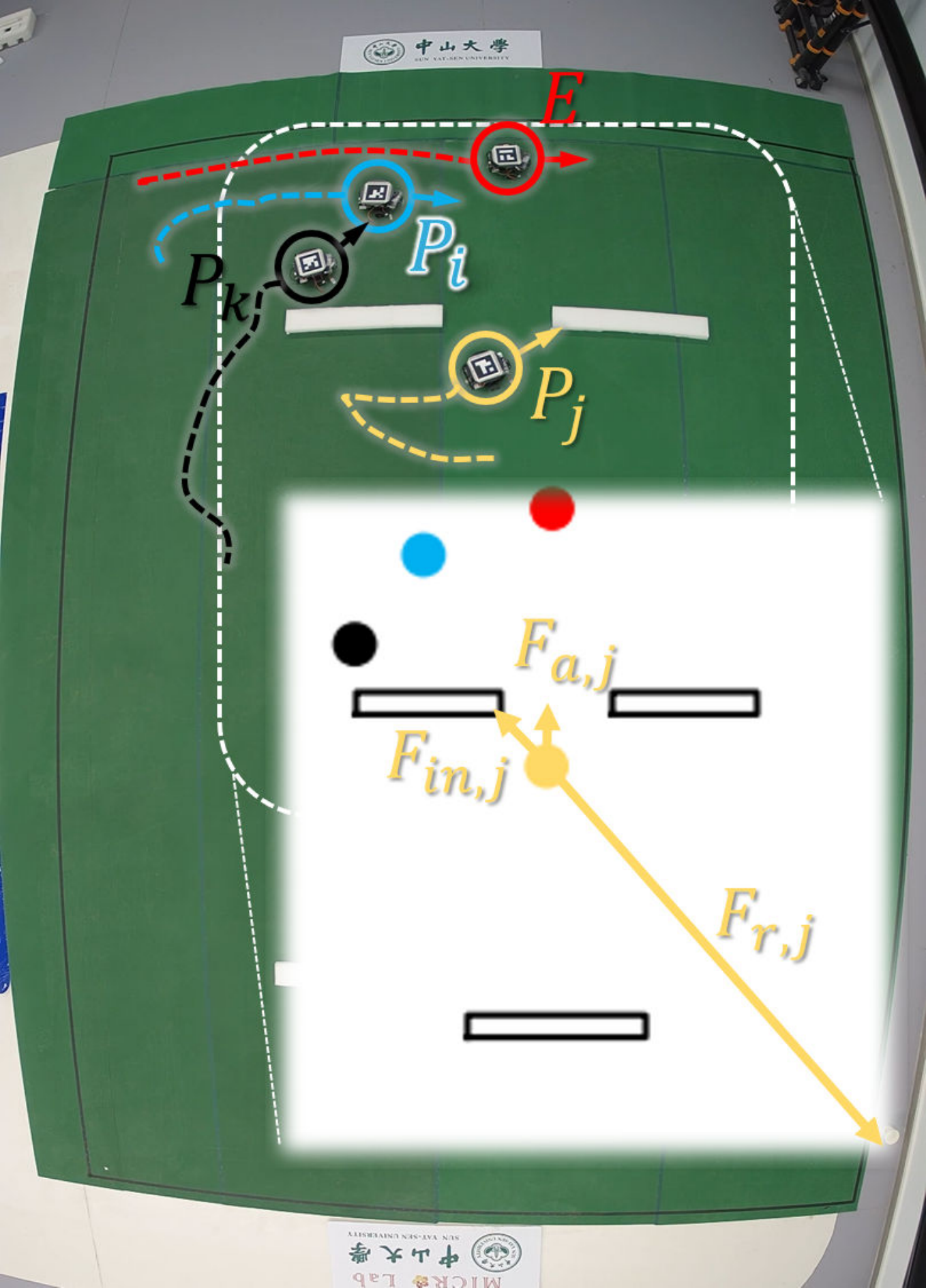}}
    \subfigure[$t=25.08$ s]{
    \includegraphics[scale=0.18]{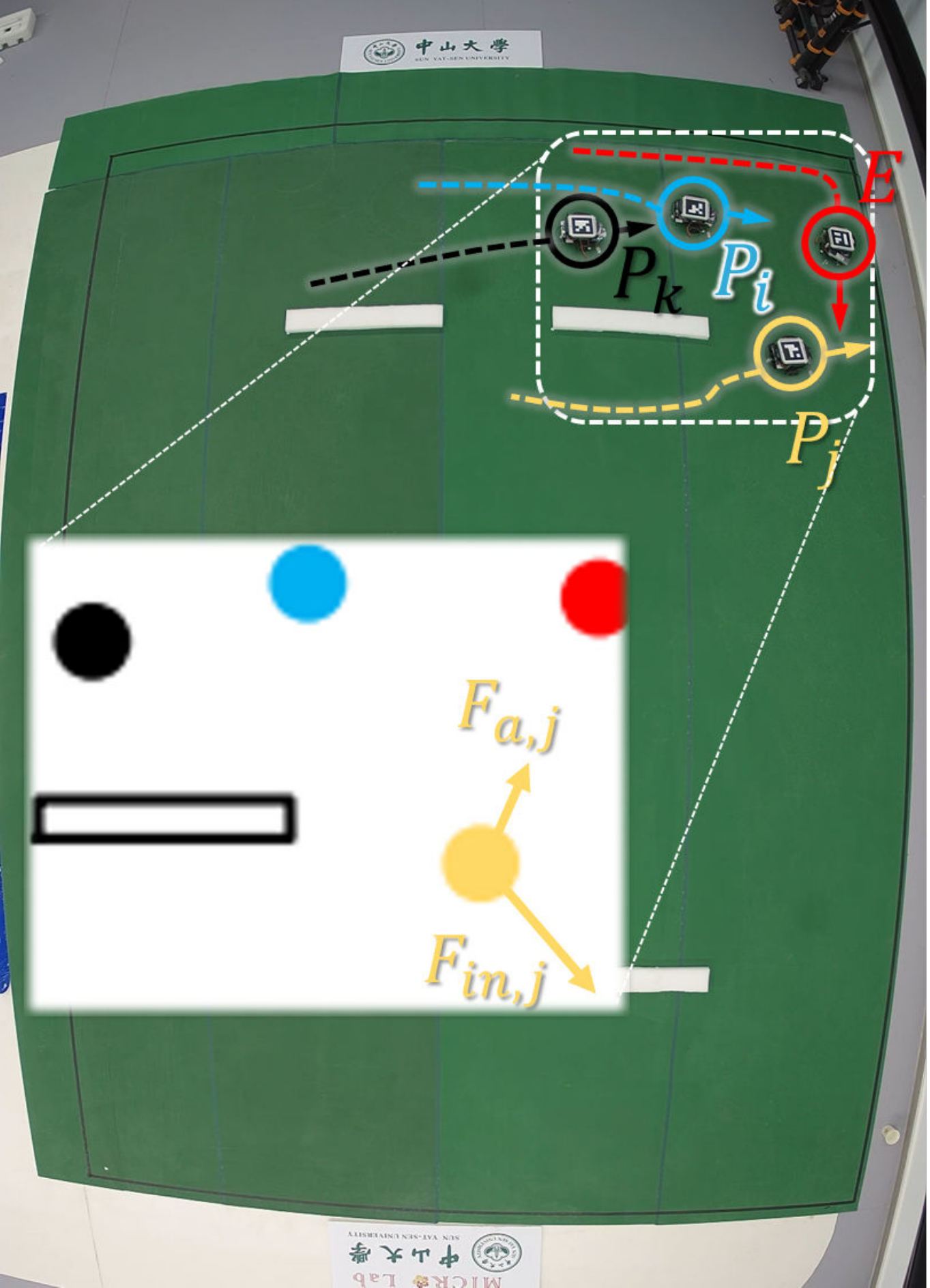}}
    \subfigure[$t=27.20$ s]{
    \includegraphics[scale=0.18]{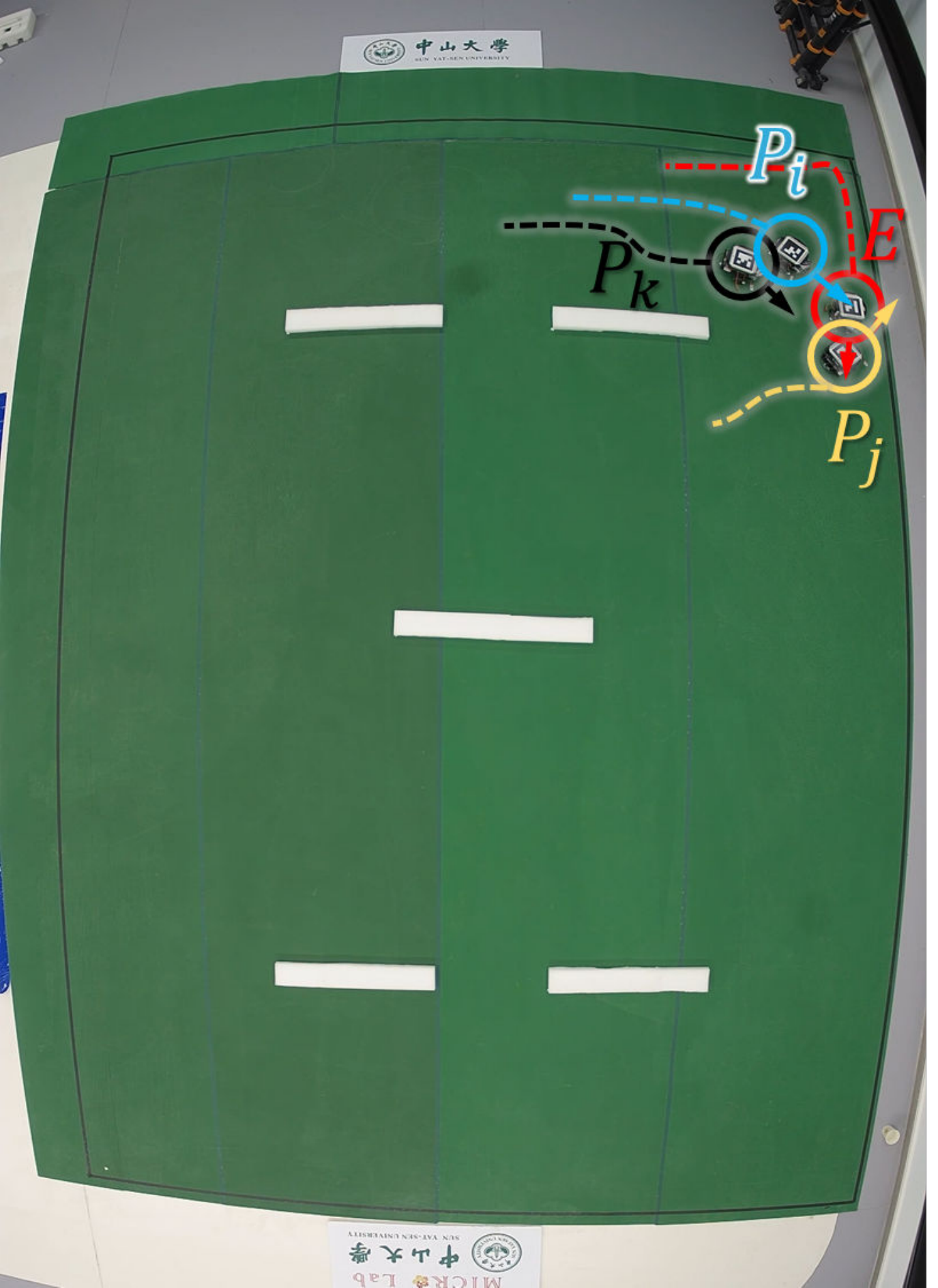}}
    \caption{The adaptive pursuit strategy according to the relative position in the team.}
    \label{fig:relative position}
\end{figure}
In the third set of snapshots shown in Figure \ref{fig:relative position}, $P_j$ could have approached the evader from the gap between obstacles, but instead it moves to the right of the map to flank the evader by selecting a large $\eta$ (Fig. \ref{fig:relative position}(a)-(b)). Similarly, $P_j$ in Fig. \ref{fig:relative position}(b) could choose to approach the evader directly, but it continues moving to the right to cut off the evader’s escaping route. Note that $P_k$ and $P_i$ tend to chase the evader directly while $P_j$ flanks it. This case demonstrates the learned policy is adaptive to the relative position in the team. Although pursuers share the same pursuit policy, they can play different roles spontaneously according to their local observations.

%%%%%%%%%%%%%%%%%%%%%%%
\section{Concluding remarks}\label{sec:conclusion}
A novel decentralized cooperative pursuit algorithm called DACOOP is presented in this paper by integrating deep RL with APF. The DACOOP algorithm is able to fuse advantages from both learning-based methods and rule-based designs to improve the pursuit performance significantly. Simulation results showed that DACOOP achieves a higher success rate in the cooperative pursuit problem than APF. In addition, it also outperforms D3QN in terms of data efficiency and generalization. The learned policy was deployed in real-world differential wheeled robots using a direct sim-to-real transfer. We demonstrated that the learned policy was adaptive to different situations. Since the pursuers can not observe the evader all the time in the real world, an efficient search of the arena without a target signal will be our future work.

%%%%%%%%%%%%%%%%%%%%%%%%

%%%%%%%%%%%%%%%%%%%%%%%%%%%%%%%%%%%%%%%%%%%%%%%%%
\bibliography{references}

\begin{thebibliography}{10}
\providecommand{\url}[1]{#1}
\csname url@samestyle\endcsname
\providecommand{\newblock}{\relax}
\providecommand{\bibinfo}[2]{#2}
\providecommand{\BIBentrySTDinterwordspacing}{\spaceskip=0pt\relax}
\providecommand{\BIBentryALTinterwordstretchfactor}{4}
\providecommand{\BIBentryALTinterwordspacing}{\spaceskip=\fontdimen2\font plus
\BIBentryALTinterwordstretchfactor\fontdimen3\font minus
  \fontdimen4\font\relax}
\providecommand{\BIBforeignlanguage}[2]{{%
\expandafter\ifx\csname l@#1\endcsname\relax
\typeout{** WARNING: IEEEtran.bst: No hyphenation pattern has been}%
\typeout{** loaded for the language `#1'. Using the pattern for}%
\typeout{** the default language instead.}%
\else
\language=\csname l@#1\endcsname
\fi
#2}}
\providecommand{\BIBdecl}{\relax}
\BIBdecl

\bibitem{bayindir2016review}
L.~Bay{\i}nd{\i}r, ``A review of swarm robotics tasks,'' \emph{Neurocomputing},
  vol. 172, pp. 292--321, 2016.

\bibitem{castanedo2010data}
F.~Castanedo, J.~Garc{\'\i}a, M.~A. Patricio, and J.~M. Molina, ``Data fusion
  to improve trajectory tracking in a cooperative surveillance multi-agent
  architecture,'' \emph{Information Fusion}, vol.~11, no.~3, pp. 243--255,
  2010.

\bibitem{allouche2010multi}
M.~K. Allouche and A.~Boukhtouta, ``Multi-agent coordination by temporal plan
  fusion: Application to combat search and rescue,'' \emph{Information Fusion},
  vol.~11, no.~3, pp. 220--232, 2010.

\bibitem{sun2021multi}
Z.~Sun, H.~Piao, Z.~Yang, Y.~Zhao, G.~Zhan, D.~Zhou, G.~Meng, H.~Chen, X.~Chen,
  B.~Qu \emph{et~al.}, ``Multi-agent hierarchical policy gradient for air
  combat tactics emergence via self-play,'' \emph{Engineering Applications of
  Artificial Intelligence}, vol.~98, p. 104112, 2021.

\bibitem{chung2011search}
T.~H. Chung, G.~A. Hollinger, and V.~Isler, ``Search and pursuit-evasion in
  mobile robotics,'' \emph{Autonomous robots}, vol.~31, no.~4, pp. 299--316,
  2011.

\bibitem{lavalle2001visibility}
S.~M. LaValle and J.~E. Hinrichsen, ``Visibility-based pursuit-evasion: The
  case of curved environments,'' \emph{IEEE Transactions on Robotics and
  Automation}, vol.~17, no.~2, pp. 196--202, 2001.

\bibitem{gerkey2006visibility}
B.~P. Gerkey, S.~Thrun, and G.~Gordon, ``Visibility-based pursuit-evasion with
  limited field of view,'' \emph{The International Journal of Robotics
  Research}, vol.~25, no.~4, pp. 299--315, 2006.

\bibitem{shah2019multi}
K.~Shah and M.~Schwager, ``Multi-agent cooperative pursuit-evasion strategies
  under uncertainty,'' in \emph{Distributed Autonomous Robotic Systems}.\hskip
  1em plus 0.5em minus 0.4em\relax Springer, 2019, pp. 451--468.

\bibitem{long2018towards}
P.~Long, T.~Fan, X.~Liao, W.~Liu, H.~Zhang, and J.~Pan, ``Towards optimally
  decentralized multi-robot collision avoidance via deep reinforcement
  learning,'' in \emph{2018 IEEE International Conference on Robotics and
  Automation (ICRA)}.\hskip 1em plus 0.5em minus 0.4em\relax IEEE, 2018, pp.
  6252--6259.

\bibitem{gupta2017cooperative}
J.~K. Gupta, M.~Egorov, and M.~Kochenderfer, ``Cooperative multi-agent control
  using deep reinforcement learning,'' in \emph{International Conference on
  Autonomous Agents and Multiagent Systems}.\hskip 1em plus 0.5em minus
  0.4em\relax Springer, 2017, pp. 66--83.

\bibitem{huttenrauch2019deep}
M.~H{\"u}ttenrauch, S.~Adrian, G.~Neumann \emph{et~al.}, ``Deep reinforcement
  learning for swarm systems,'' \emph{Journal of Machine Learning Research},
  vol.~20, no.~54, pp. 1--31, 2019.

\bibitem{borenstein1989real}
J.~Borenstein and Y.~Koren, ``Real-time obstacle avoidance for fast mobile
  robots,'' \emph{IEEE Transactions on systems, Man, and Cybernetics}, vol.~19,
  no.~5, pp. 1179--1187, 1989.

\bibitem{yun1997wall}
X.~Yun and K.-C. Tan, ``A wall-following method for escaping local minima in
  potential field based motion planning,'' in \emph{1997 8th International
  Conference on Advanced Robotics. Proceedings. ICAR'97}.\hskip 1em plus 0.5em
  minus 0.4em\relax IEEE, 1997, pp. 421--426.

\bibitem{ge2002dynamic}
S.~S. Ge and Y.~J. Cui, ``Dynamic motion planning for mobile robots using
  potential field method,'' \emph{Autonomous robots}, vol.~13, no.~3, pp.
  207--222, 2002.

\bibitem{angelani2012collective}
L.~Angelani, ``Collective predation and escape strategies,'' \emph{Physical
  review letters}, vol. 109, no.~11, p. 118104, 2012.

\bibitem{khatib1986real}
O.~Khatib, ``Real-time obstacle avoidance for manipulators and mobile robots,''
  in \emph{Autonomous robot vehicles}.\hskip 1em plus 0.5em minus 0.4em\relax
  Springer, 1986, pp. 396--404.

\bibitem{janosov2017group}
M.~Janosov, C.~Vir{\'a}gh, G.~V{\'a}s{\'a}rhelyi, and T.~Vicsek, ``Group
  chasing tactics: how to catch a faster prey,'' \emph{New Journal of Physics},
  vol.~19, no.~5, p. 053003, 2017.

\bibitem{escobedo2014group}
R.~Escobedo, C.~Muro, L.~Spector, and R.~Coppinger, ``Group size, individual
  role differentiation and effectiveness of cooperation in a homogeneous group
  of hunters,'' \emph{Journal of the Royal Society Interface}, vol.~11, no.~95,
  p. 20140204, 2014.

\bibitem{muro2011wolf}
C.~Muro, R.~Escobedo, L.~Spector, and R.~Coppinger, ``Wolf-pack (canis lupus)
  hunting strategies emerge from simple rules in computational simulations,''
  \emph{Behavioural processes}, vol.~88, no.~3, pp. 192--197, 2011.

\bibitem{fang2020cooperative}
X.~Fang, C.~Wang, L.~Xie, and J.~Chen, ``Cooperative pursuit with multi-pursuer
  and one faster free-moving evader,'' \emph{IEEE transactions on cybernetics},
  2020.

\bibitem{sutton2018reinforcement}
R.~S. Sutton and A.~G. Barto, \emph{Reinforcement learning: An
  introduction}.\hskip 1em plus 0.5em minus 0.4em\relax MIT press, 2018.

\bibitem{silver2021reward}
D.~Silver, S.~Singh, D.~Precup, and R.~S. Sutton, ``Reward is enough,''
  \emph{Artificial Intelligence}, p. 103535, 2021.

\bibitem{wang2020cooperative}
Y.~Wang, L.~Dong, and C.~Sun, ``Cooperative control for multi-player
  pursuit-evasion games with reinforcement learning,'' \emph{Neurocomputing},
  vol. 412, pp. 101--114, 2020.

\bibitem{nair2003taming}
R.~Nair, M.~Tambe, M.~Yokoo, D.~Pynadath, and S.~Marsella, ``Taming
  decentralized pomdps: Towards efficient policy computation for multiagent
  settings,'' in \emph{IJCAI}, vol.~3.\hskip 1em plus 0.5em minus 0.4em\relax
  Citeseer, 2003, pp. 705--711.

\bibitem{de2021decentralized}
C.~de~Souza, R.~Newbury, A.~Cosgun, P.~Castillo, B.~Vidolov, and D.~Kuli{\'c},
  ``Decentralized multi-agent pursuit using deep reinforcement learning,''
  \emph{IEEE Robotics and Automation Letters}, vol.~6, no.~3, pp. 4552--4559,
  2021.

\bibitem{rlblogpost}
A.~Irpan, ``Deep reinforcement learning doesn't work yet,''
  \url{https://www.alexirpan.com/2018/02/14/rl-hard.html}, 2018.

\bibitem{mnih2015human}
V.~Mnih, K.~Kavukcuoglu, D.~Silver, A.~A. Rusu, J.~Veness, M.~G. Bellemare,
  A.~Graves, M.~Riedmiller, A.~K. Fidjeland, G.~Ostrovski \emph{et~al.},
  ``Human-level control through deep reinforcement learning,'' \emph{nature},
  vol. 518, no. 7540, pp. 529--533, 2015.

\bibitem{koren1991potential}
Y.~Koren and J.~Borenstein, ``Potential field methods and their inherent
  limitations for mobile robot navigation,'' in \emph{Proceedings. 1991 IEEE
  International Conference on Robotics and Automation}.\hskip 1em plus 0.5em
  minus 0.4em\relax IEEE, 1991, pp. 1398--1404.

\bibitem{ge2000new}
S.~S. Ge and Y.~J. Cui, ``New potential functions for mobile robot path
  planning,'' \emph{IEEE Transactions on robotics and automation}, vol.~16,
  no.~5, pp. 615--620, 2000.

\bibitem{schaul2015prioritized}
T.~Schaul, J.~Quan, I.~Antonoglou, and D.~Silver, ``Prioritized experience
  replay,'' \emph{arXiv preprint arXiv:1511.05952}, 2015.

\bibitem{wang2016dueling}
Z.~Wang, T.~Schaul, M.~Hessel, H.~Hasselt, M.~Lanctot, and N.~Freitas,
  ``Dueling network architectures for deep reinforcement learning,'' in
  \emph{International conference on machine learning}.\hskip 1em plus 0.5em
  minus 0.4em\relax PMLR, 2016, pp. 1995--2003.

\bibitem{ng1999policy}
A.~Y. Ng, D.~Harada, and S.~J. Russell, ``Policy invariance under reward
  transformations: Theory and application to reward shaping,'' in
  \emph{Proceedings of the Sixteenth International Conference on Machine
  Learning}, 1999, pp. 278--287.

\bibitem{matignon2012independent}
L.~Matignon, G.~J. Laurent, and N.~Le~Fort-Piat, ``Independent reinforcement
  learners in cooperative markov games: a survey regarding coordination
  problems,'' \emph{The Knowledge Engineering Review}, vol.~27, no.~1, pp.
  1--31, 2012.

\bibitem{kingma2014adam}
D.~P. Kingma and J.~Ba, ``Adam: A method for stochastic optimization,''
  \emph{arXiv preprint arXiv:1412.6980}, 2014.

\end{thebibliography}
\bibliographystyle{IEEEtran}
\end{document}